\documentclass[10pt,final, twocolumn]{IEEEtran}
\usepackage[utf8]{inputenc}
\def\BibTeX{{\rm B\kern-.05em{\sc i\kern-.025em b}\kern-.08em
    T\kern-.1667em\lower.7ex\hbox{E}\kern-.125emX}}

\usepackage{amsmath,amssymb}
\usepackage{algorithm}
\usepackage{algpseudocode}
\usepackage{graphicx}
\usepackage{color}
\usepackage{xcolor}
\usepackage{cite}
\usepackage[caption=false]{subfig}
\usepackage{comment}
\usepackage{mathtools}
\usepackage{tabulary}
\usepackage{acronym}
\usepackage{upgreek}
\usepackage{algpseudocode}
\usepackage{placeins}
\usepackage[super]{nth}
\usepackage{dsfont}
\usepackage{hyperref}
\usepackage{array}
\usepackage{soul}
\usepackage{dsfont}
\usepackage{cancel}
\usepackage{multirow,flushend}

\definecolor{darkgreen}{rgb}{0.0, 0.5, 0.0}

\newcommand{\myvec}[1]{\boldsymbol{#1}}

\usepackage[acronym]{glossaries}
\newacronym{RIS}{RIS}{Reconfigurable Intelligent Surface}
\newacronym{SIM}{SIM}{Stacked Intelligent Metasurfaces}
\newacronym{DNN}{DNN}{Deep artificial Neural Network}
\newacronym{EI}{EI}{Edge Inference}
\newacronym{TOC}{GOC}{Goal-Oriented Communications}
\newacronym{TX}{TX}{Transmitter}
\newacronym{RX}{RX}{Receiver}
\newacronym{UE}{UE}{User Equipment}
\newacronym{BS}{BS}{Base Station}
\newacronym{LoS}{LoS}{Line-of-Sight}
\newacronym{NLoS}{NLoS}{Non Line-of-Sight}
\newacronym{CNN}{CNN}{Convolutional Neural Network}
\newacronym{FFNN}{FFNN}{Feed-Forward Neural Network}
\newacronym{ReLU}{ReLU}{Rectified Linear Unit}
\newacronym{JSCC}{JSCC}{Joint Source Channel Coding}
\newacronym{E2E}{E2E}{End-to-End}
\newacronym{D2D}{D2D}{Device-to-Device}
\newacronym{DeepSC}{DeepSC}{Deep Semantic Communications}
\newacronym{SISO}{SISO}{Single Input Single Output}
\newacronym{MISO}{MISO}{Multiple Input Single Output}
\newacronym{MIMO}{MIMO}{Multiple-Input Multiple-Output}
\newacronym{OAC}{OAC}{Over-the-Air Computation}
\newacronym{MSE}{MSE}{Mean Squared Error}
\newacronym{SOTA}{SotA}{State of the Art}
\newacronym{SGD}{SGD}{Stochastic Gradient Descent}
\newacronym{IoT}{IoT}{Internet of Things}
\newacronym{AE}{AE}{Auto-Encoder}
\newacronym{PDF}{PDF}{Probability Density Function}
\newacronym{AWGN}{AWGN}{Additive White Gaussian Noise}
\newacronym{CSI}{CSI}{Channel State Information}
\newacronym{CFR}{CFR}{Channel Frequency Response}
\newacronym{ISAC}{ISAC}{Integrated Sensing and Communications}
\newacronym{iid}{i.i.d.}{independent and identically distributed}
\newacronym{MINN}{MINN}{Metasurfaces-Integrated Neural Network}
\newacronym{MS}{MS}{Meta-Surface}
\newacronym{CE}{CE}{Cross Entropy}
\newacronym{HRIS}{HRIS}{Hybrid RIS}
\newacronym{RF}{RF}{Radio-Frequency}
\newacronym{VAE}{VAE}{Variational Auto-Encoder}
\newacronym{PSK}{PSK}{Phase Shift Keying}
\newacronym{SVD}{SVD}{Singular Value Decomposition}
\newacronym{MLP}{MLP}{Multi Layer Perceptron}
\newacronym{SNR}{SNR}{Signal-to-Noise Ratio}
\newacronym{MI}{MI}{Mutual Information}
\newacronym{MEC}{MEC}{Multi-access Edge Computing}
\newacronym{WMMSE}{WMMSE}{Weighted Minimum Mean Square Error}

\glsdisablehyper

\title{Over-the-Air Edge Inference via End-to-End Metasurfaces-Integrated Artificial Neural Networks}

\author{Kyriakos Stylianopoulos,~\IEEEmembership{Graduate~Student~Member,~IEEE,} Paolo~Di~Lorenzo,~\IEEEmembership{Senior~Member,~IEEE,}\\ and George~C.~Alexandropoulos,~\IEEEmembership{Senior~Member,~IEEE}
\thanks{This work has been supported by the Smart Networks and Services
Joint Undertaking projects TERRAMETA, 6G-DISAC, and 6G-GOALS under
the European Union's Horizon Europe research and innovation programme
under Grant Agreement numbers 101097101, 101139130, and 101139232 respectively.
TERRAMETA also includes top-up funding by UK Research and Innovation under the UK government’s Horizon Europe funding guarantee.}
\thanks{K. Stylianopoulos and G.~C.~Alexandropoulos are with the Department of Informatics and Telecommunications, National and Kapodistrian University of Athens, 16122 Athens, Greece
(e-mails: \{kstylianop, alexandg\}@di.uoa.gr).}
\thanks{P.~Di~Lorenzo is with the Department of Information Engineering, Electronics, and Telecommunications, Sapienza University, Italy and 
CNIT, Italy (e-mail: paolo.dilorenzo@uniroma1.it).}
}

\begin{document}
\maketitle

\thispagestyle{empty}
\pagestyle{empty}

\begin{abstract}
In the Edge Inference (EI) paradigm, where a Deep Neural Network (DNN) is split across the transceivers to wirelessly communicate goal-defined features in solving a computational task, the wireless medium has been commonly treated as a source of noise. In this paper, motivated by the emerging technologies of Reconfigurable Intelligent Surfaces (RISs) and Stacked Intelligent Metasurfaces (SIM) that offer programmable propagation of wireless signals, either through controllable reflections or diffractions, we optimize the RIS/SIM-enabled smart wireless environment as a means of over-the-air computing, resembling the operations of DNN layers. We propose a framework of Metasurfaces-Integrated Neural Networks (MINNs) for EI, presenting its modeling, training through a backpropagation variation for fading channels, and deployment aspects. The overall end-to-end DNN architecture is general enough to admit RIS and SIM devices, through controllable reconfiguration before each transmission or fixed configurations after training, while both channel-aware and channel-agnostic transceivers are considered. Our numerical evaluation showcases metasurfaces to be instrumental in performing image classification under link budgets that impede conventional communications or metasurface-free systems. It is demonstrated that our MINN framework can significantly simplify EI requirements, achieving near-optimal performance with $50$~dB lower testing signal-to-noise ratio compared to training, even without transceiver channel knowledge.
\end{abstract}
\vspace{-0.18cm}
\begin{IEEEkeywords}
Edge learning, reconfigurable intelligent surface, stacked intelligent metasurfaces, goal-oriented communications, deep learning, over-the-air computing.
\end{IEEEkeywords}

\section{Introduction}
In emerging \gls{D2D} and \gls{IoT} networks, where distributed devices are expected to support various functionalities, such as \gls{ISAC}~\cite{6G-DISAC}, stringent energy efficiency requirements impose limitations on their communication and computation capabilities.
Various sub-networks are envisioned to enable a wide range of high-level applications and services, such as digital twinning through object recognition and computational imaging, or indoor positioning~\cite{6G-IA-Vision}, each tied up with a diverse set of requirements to be fulfilled.
As a result, it is crucial for the \gls{D2D} system design to take into consideration the underlying application. To this end, devising novel communication stacks tailored to the application, that break the traditional network layer taxonomy, comes with reduced implementation overheads.

\Gls{TOC}~\cite{DMB23} is a framework that is gaining popularity in \gls{D2D} systems, since transmissions of only the necessary goal-specific information need to take place, reducing messaging overheads and simplifying system architecture~\cite{6G-GOALS}.
In fact, when performing \gls{EI} tasks, where the \gls{RX} wishes to obtain only an estimated label of the data the \gls{TX} actually observes and transmits, semantic and \gls{TOC} approaches that split the layers of a \gls{DNN} at the endpoints provide significant benefits in terms of communication requirements~\cite{MLB21}. According to those approaches, low-dimensional feature vectors, that are outputs of intermediate \gls{DNN} layers, are transmitted over the channel~\cite{PBD22_bottleneck, DeepOAC, DeepSC}.
The motivation behind such practices is that \gls{TOC} deviates from the standard Shannon-type communications~\cite{shannon}, in that the goal of the system is an arbitrary computational target function, rather than objectives derived from the \gls{MI} that call for bit-wise reconstruction of the input data. In this way, \gls{DNN}s are employed, similar to the \gls{JSCC} paradigm, and trained to capture the joint or conditional data-channel-target distributions.

Current literature has adopted \gls{TOC} and \gls{EI} for a wide variety of problems and wireless systems (see~\cite{PBD22_bottleneck, DeepSC, JGM21_Image_Retrieval, LWM23} and the references therein), however, a common practice across those works is to treat the wireless environment as a source of noise, whose effects need to be negated at the \gls{RX} side.
The rapid developments of \gls{MS} technologies for precise \gls{RF} domain control open the potential for the wireless propagation medium to be dynamically reconfigurable via reflective \glspl{RIS}~\cite{BAL24} or diffractive \gls{SIM}~\cite{AXN23} with low operational costs.
Such \gls{MS}s have been incorporated in \gls{TOC} and semantic systems to reduce the transceiver hardware complexity~\cite{GJZ24_SIM_TOC} or to enhance the system's rate~\cite{HMC24_Mattias_RIS}.
Under another research direction, \gls{DNN} implementations entirely in the \gls{RF} domain, capitalizing on \gls{MS}-based solutions, have been devised for controlled laboratory environments~\cite{XYN18, LML23, CHQ24}.

\begin{figure}[t]
    \centering
    \includegraphics[width=0.8\linewidth]{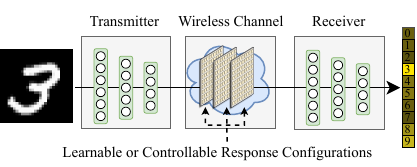}
    \vspace{-0.3cm}
    \caption{A Metasurfaces-Integrated artificial Neural Network (MINN) performing Edge Inference (EI) by controlling the wireless propagation channel, which is treated as one or more hidden network layers.}
    \label{fig:arch}
    \vspace{-0.3cm}
\end{figure}
In this paper, motivated by the elaborate wireless medium control offered by MS technologies~\cite{BAL24,AXN23}, in conjunction with their analog computational capabilities, we design \gls{MS}-controllable wireless channels that perform \gls{OAC}, under which the comprising metamaterials are treated as hidden artificial neurons that control the wireless medium to perform multi-layer non-linear signal processing toward solving \gls{EI} tasks, as illustrated in Fig.~\ref{fig:arch}. The main contributions of this paper are summarized as follows.
\begin{enumerate}
    \item We present a novel generic \gls{E2E} \gls{DNN} framework, titled \gls{MINN}, that admits variations of controllable \glspl{MS} or \glspl{MS} of trainable, yet fixed response configurations, of either \glspl{RIS} or \gls{SIM} and with or without channel knowledge. We detail the framework's modeling, backpropagation-based training, and deployment from both a theoretical and a system architecture perspectives.
    \item We elaborate on the critical role of reconfiguration of wireless channels as a degree of freedom in optimizing \gls{DNN}-based \gls{EI} models that capitalize on programmable \gls{OAC}, instead of treating the wireless propagation environment as a source of noise. To this end, learning \glspl{RIS} and \gls{SIM} configurations promises substantial  potential. 
    \item We present an extensive numerical evaluation of the proposed \gls{MINN} framework on an image classification task, which compared with conventional communication systems, as well as a baseline in the absence of an \gls{MS}, showcases that \gls{MS}-enabled \gls{OAC} allows for successful \gls{EI} with much lower transmit power requirements, even without channel knowledge at the communication ends.
\end{enumerate}

The remainder of the paper is structured as follows. Section~\ref{sec:background} details the relevant pieces of theory regarding \gls{EI} as well as the possible improvements brought by controlling the wireless environment, while Section~\ref{sec:literature} provides a comprehensive review of the relevant literature. Section~\ref{sec:system-model} includes the models used for the considered \gls{MS} technologies: \gls{RIS} and \gls{SIM}, as well as the model for the received signal for both cases. The proposed \gls{MINN} architecture for \gls{EI} is presented in Section~\ref{sec:architecture}, 
whereas Section~\ref{sec:training-deployment} details its training procedure and discusses the deployment and network considerations. Our numerical investigations are presented in Section~\ref{sec:results}.
Finally, Section~\ref{sec:conclusion} includes the paper's concluding remarks.

\textbf{Notation:}
Vectors, matrices, and sets are expressed in lowercase bold (e.g., $\myvec{x}$), uppercase bold (e.g., $\myvec{X}$), and uppercase calligraphic typefaces (e.g., $\mathcal{X}$ or $\boldsymbol{\mathcal{X}}$), respectively.
Apart from $\myvec{X}^{\ast}$, $\myvec{X}^\dagger$, and $\myvec{X}^\top$ that denote the conjugate, conjugate transpose, and transpose of $\myvec{X}$,
superscripts and subscripts are used to denote different versions of variables or enumeration over collections of variables depending on the context.
$[\myvec{x}]_i$ and $[\myvec{X}]_{i,j}$ are used to denote the $i$-th element of $\myvec{x}$ and the $(i,j)$-th element of $\myvec{X}$, respectively.
We use variations of the notation $f_{\myvec{w}}(\cdot)$ to represent neural network functions that are parameterized by their weight matrix $\myvec{w}$, noting that $f$ may be seen either as a function of its arbitrary input variables (during inference) or as a function of $\myvec{w}$ (during optimization).
$|\mathcal{X}|$ represents the cardinality of the set $\mathcal{X}$, ${\rm diag(\myvec{x})}$ creates a square matrix with the elements of $\myvec{x}$ placed along its main diagonal, and ${\rm vec}(\myvec{X})$ transforms $\myvec{X}$ in a column vector in a row-by-row fashion. $\otimes$ denotes the Kronecker product, $\{a,b\}$ expresses a set or collection containing $a$ and $b$, while $\mathds{1}_{\texttt{cond}}$ is the indicator function equaling to $1$ if condition $\texttt{cond}$ holds, otherwise to $0$. Finally, $\mathbb{E}_{\myvec{X}}[\cdot]$ is the expectation operator with respect to the distribution of the random $\myvec{X}$, and $\jmath\triangleq\sqrt{-1}$.

\section{Prerequisites}\label{sec:background}

\subsection{Probabilistic Inference}\label{sec:inference-theory}
Given an input observation $\myvec{x}$, the objective of an inference procedure is to compute and output an associated target  
attribute value  $\myvec{o} = l(\myvec{x})$.
The mapping function $l(\myvec{x})$ is considered unknown and intractable to express analytically, therefore, inference involves approximating this relationship through examples of $(\myvec{x}, \myvec{o})$ tuples.
From a probabilistic perspective, one may fit the conditional \gls{PDF} of the target $p(\myvec{o}|\myvec{x})$ on the available data.
In typical settings, point estimates are only required, thus, the problem reduces in predicting the most likely value of $\myvec{o}$ for a given observation.
In the machine learning regime, the previous distribution (or its point estimates) can be approximated by a $\myvec{w}$-parameterized model $\hat{\myvec{o}} \triangleq f_{\myvec{w}}(\myvec{x})$ that outputs its prediction of the target value $\hat{\myvec{o}}$ for a given input.
Consecutively, to solve the inference problem, the parameter values $\myvec{w}$ need to be optimized.
This is achieved by collecting a data set of training tuples $\mathcal{D} \triangleq  \{ (\myvec{x}_i, \myvec{o}_i) \}_{i=1}^{|\mathcal{D}|}$, and then minimizing an amortized cost function over the training instances:
\begin{align}\label{eq:cost-function}
    & J(\myvec{w}) \triangleq \frac{1}{|\mathcal{D}|} \sum_{i=1}^{|\mathcal{D}|} J_{\rm str}(\myvec{o}_i, \hat{\myvec{o}}_i), \quad\text{where } \hat{\myvec{o}}_i = f_{\myvec{w}}(\myvec{x}_i).
\end{align} 
\par
The per-instance cost function values $J_{\rm str}(\myvec{o}_i, \hat{\myvec{o}}_i)$'s quantify the error in the model's predictions and they often have a probabilistic interpretation.
For instance, in {\em classification} settings where each observation belongs to one of $d_{\rm cl}$ classes indexed by the natural number $c$, the target value is defined via one-hot encoding as $\myvec{o} = [\mathds{1}_{c=1}, \mathds{1}_{c=2}, \dots, \mathds{1}_{c=d_{\rm cl}}]^\top \in \mathbb{N}^{d_{\rm cl} \times 1}$. The \gls{CE} loss function is defined as follows:
\begin{equation}\label{eq:cross-entropy}
    J_{\rm CE}(\myvec{o}_i, \hat{\myvec{o}}_i) \triangleq - \sum_{j=1}^{d_{\rm cl}} [\myvec{o}_i]_j \log [\hat{\myvec{o}}_i]_j.
\end{equation}
Minimizing \eqref{eq:cross-entropy} over $\mathcal{D}$ is equivalent to performing maximum likelihood estimation of $\myvec{w}$ on $p(\myvec{o}|\myvec{x})$ under the assumption that this PDF is a multivariate Bernoulli distribution.
Similarly, in {\em regression} tasks, where $\myvec{o}, \hat{\myvec{o}} \in \mathbb{R}^{d_{\rm out} \times 1}$, the \gls{MSE} metric, defined as $1/d_{\rm out} \|\myvec{o} - \hat{\myvec{o}} \|^2$, implies that $p(\myvec{o}|\myvec{x})$ is a Gaussian conditional PDF.

\subsection{Artificial Neural Networks}\label{sec:NN-theory}
While a wide range of parameterized families of functions is available to use as $f_{\myvec{w}}(\cdot)$, the current state of the art considers \gls{DNN} models, capitalizing on their diverse benefits including high expressivity, diverse selection of architectural components for different tasks, and parallelizable computations that offer real-time computational cost during inference on pertinent hardware. Mathematically, let a neural network be expressed as a composition of $L$ layers such that:
\begin{equation}\label{eq:neuralnet}
    f_{\myvec{w}}(\myvec{x}) \triangleq f^{L}_{\myvec{w}^L}\left( f^{L-1}_{\myvec{w}^{L-1}}\left(\ldots f^{1}_{\myvec{w}^1}( \myvec{x} ) \ldots \right)\right),
\end{equation}
so that the $l$-th layer ($l=1,2,\ldots,L$) is parameterized by $\myvec{w}_l$, and its output $\bar{\myvec{o}}^l$ constitutes the input to the $(l+1)$-th layer; this process can be represented recursively by $\bar{\myvec{o}}^l \triangleq f^{l}_{\myvec{w}^l}( \bar{\myvec{o}}^{(l-1)} )$, where we have set $\bar{\myvec{o}}^0 = \myvec{x}$.
For convenience, let us also define the vector $\myvec{w} \triangleq [\myvec{w}^{\top}_1, \myvec{w}^\top_2, \dots, \myvec{w}^\top_L]^\top$.
\par
We leave aside precise definitions of individual layer functions, as there has been an impressive body of work focusing on layers that perform data-specific computations and capture high-level patterns in data sets \cite{deep_learning_book}.
Nevertheless, we highlight the fact that each $f^l_{\myvec{w}_l}(\cdot)$ is demanded to be a non-linear function, and, in fact, it is often requested to be discriminatory or sigmoidal~\cite{Cybenko89}.
Those properties guarantee that artificial neural networks with at least two layers (of potentially infinite width) are {\em universal approximators} and can, therefore, be used to approximate any arbitrary $\myvec{o} = l(\myvec{x})$ mapping, i.e., $f_{\myvec{w}}(\myvec{x})\cong l(\myvec{x})$~\cite{Cybenko89}.
\par
Besides theoretical guarantees, the problem of obtaining $\myvec{w}$ values that perform successful inference can be efficiently solved by substituting the neural network expression from \eqref{eq:neuralnet} into \eqref{eq:cross-entropy} (using classification as an example), and subsequently into \eqref{eq:cost-function}.
The latter may be solved through one of the many variations of the \gls{SGD} approach by computing the gradients $\partial J(\myvec{w})/\partial\myvec{w}$, and propagating them through the layers of $f_{\myvec{w}}(\myvec{x})$ by taking advantage of the chain rule; this leads to the celebrated {\em backpropagation} algorithm \cite{Rumelhart86} that constitutes the backbone of deep learning.

\subsection{Edge Inference (EI)}\label{sec:EI-theory}
The rising field of \gls{EI} considers the implementation and training of inference tasks over a wireless communication network.
In that regard, consider an uplink setup where a multi-antenna \gls{TX} observes $\myvec{x}$ and wishes to convey its estimation $\hat{\myvec{o}}$ of the target value $\myvec{o}$ to an \gls{RX}. At first observation, this task is apparently straightforward to implement within the existing frameworks of machine learning and wireless communications under the following two paradigm options:

\subsubsection{``Infer-then-transmit''}
The {\rm TX} may first compute $\hat{\myvec{o}} = f_{\myvec{w}}(\myvec{x})$ and then transmit $\hat{\myvec{o}}$ over the wireless medium upon performing source coding (i.e., data compression) and channel coding (i.e., modulation and beamforming) to derive the transmission signal $\myvec{s}$.
The above coding computations ensure that a satisfactory communication rate is achievable by the system so that $\hat{\myvec{o}}$ may be reconstructed at the \gls{RX}'s side via decoding the received signal and decompressing the data.
In modern high-complexity wireless systems, those operations require their own optimization procedures and necessitate additional computational costs as well as channel state information knowledge.
For most inference tasks, the target values are of a much smaller dimension than the observations, as a result, this option includes small rate requirements, but comes at a computational cost on the \gls{TX}'s side, as the device must be endowed with hardware capable of executing \gls{DNN} computations locally.
Since \gls{EI} tasks are envisioned specifically for cases where \gls{IoT} or other lightweight devices with low complexity and minute power consumption sending messages to a collection/fusion center, the assumption of computational capabilities for local \gls{DNN}  inference is rather optimistic.

\subsubsection{``Transmit-then-infer''}
This converse approach is also possible: the \gls{TX} performs source and channel coding but no \gls{DNN}-based \gls{EI} computations, so that the original observation $\myvec{x}$ is transmitted instead.
The \gls{RX} performs decoding to obtain the data point, which is then fed to its local $f_{\myvec{w}}(\cdot)$ to perform inference.
While in uplink settings, it is reasonable to assume that the \gls{RX} has sufficient power and hardware capabilities to support a \gls{DNN}, the rate required for transmitting the original observation may impose high link budget demands that are not readily facilitated.

A compromising solution to the above may be obtainable by exploiting the sequential nature of the \gls{DNN} structure appearing in~\eqref{eq:neuralnet}~\cite{dnn-splitting}. To this end, the following EI option is possible:
\subsubsection{``Infer-while-transmitting'' (DNN splitting)}
The intermediate representations $\bar{\myvec{o}}^l$ $\forall l=1,2\ldots,L-1$ can be of arbitrary dimensions, and it is not uncommon to devise architectures with one or more small-sized intermediate layers. In fact, various deep learning models, such as auto-encoders \cite{VAE} and U-Net \cite{U_Net}, are designed specifically to contain such bottleneck layers as a form of compression to keep only relevant information.
From that perspective, one may choose to split the first $L'<L$ layers of the \gls{DNN} to reside at the \gls{TX}, so that $\bar{\myvec{o}}^{L'}$ is transmitted over the network and, then, pass to the $(L'+1)$-th up to the $L$-th layer in sequence at the \gls{RX}.

Evidently, the latter paradigm of \gls{EI} is the most flexible, since one has the option to balance trade-offs between computation and communication resources between the TX and RX.
In the remainder of this paper, we will be adopting this paradigm of \gls{EI} as the default case and provide further elaboration and extensions, revisiting the two extreme previous cases as baselines in our numerical comparisons.

\subsection{Computational Considerations}\label{sec:computation-of-EI}
When performing DNN splitting over a wireless channel with realistic characteristics (i.e., large- and small-scale fading, as well as \gls{AWGN}), the transmitted output $\bar{\myvec{o}}^{L'}$ of the $L'$-th DNN layer will be distorted when arriving at the \gls{RX}. By representing the channel state with an abstract random variable $\boldsymbol{\mathcal{H}}$, one needs to minimize the same $\myvec{w}$-parameterized objective function as before, while accounting for the stochastic nature of the wireless environment (i.e., with respect to $\boldsymbol{\mathcal{H}}$'s distribution), i.e., solve:
\begin{equation*}
    \mathcal{OP}_{\rm EI}: \min_{\myvec{w}} \mathbb{E}_{\boldsymbol{\mathcal{H}}}[J(\myvec{w})],
\end{equation*}
where each instantaneous channel realization affects the value of the objected function by distorting the value of the transmitted $\bar{\myvec{o}}^{L'}$.
The precise definition of $\boldsymbol{\mathcal{H}}$ will be given in Section~\ref{sec:system-model}, while the considered fading distributions are discussed during the numerical evaluation in Section~\ref{sec:results}.

Assuming the wireless channel has sufficient capacity, optimizing both endpoints to perform source and channel encoding and decoding, under the standard paradigm of wireless communications, will indeed nullify the distortion on the received version of $\bar{\myvec{o}}^{L'}$.
The decoding output may then passed to the $(L'+1)$-layer of the network as normal; note that the presence of the channel is effectively {\em hidden} from the neural network's perspective.
This approach aligns more with the standard practices of both the wireless communications and machine learning paradigms, as each problem is treated individually and has shown to indeed exhibit satisfactory results \cite{DeepSC, HMC24_Mattias_RIS}. However, the above practice of optimizing the system for reconstruction of the received signal may result in higher computational overheads under the following considerations:
\begin{enumerate}
    \item Input reconstruction is not the objective of \gls{EI}, rather the computation of an arbitrary function of it.
    Under this point of view, allocating computational resources in reconstructing intermediate variables is not always the most efficient way of solving the problem.
    In fact, one may regard \gls{TOC} as a particular instance of lossy compression between the (unseen) target value $\myvec{o}$ and its estimation $\hat{\myvec{o}}$, where $J_{\rm str}(\myvec{o}, \hat{\myvec{o}})$ plays the role of the distortion metric.     From an information-theoretic perspective, variations of the distortion-rate functions may then be studied, as proposed by \cite{SK22_Rate_Distorion}, indicating that the channel rate required to transmit intermediate variables, while ensuring a desired error threshold, is typically lower than the actual channel capacity which assumes reconstruction with arbitrary small error probabilities.
    \item From an engineering perspective, since the received signal is to be fed to subsequent neural network layers, perfect reconstruction may not be required.
    The employed neural network architectures are commonly designed to account for noisy inputs in light of the inherent stochasticity of inference problems.
    Besides, manually imputing noise to activations of intermediate layers, both during training \cite{Dropout, GMD16_Noisy_activations} and inference \cite{Variational_Dropout, MC_Dropout}, is known to enhance the model's regularization and uncertainty estimation properties. 
    \item Finally, the wireless channel, which can be regarded as a (naturally stochastic) function over the transmitted data, imposes its own computations. While this function is not, in general, controllable by the E2E system, \gls{OAC} approaches leverage the superimposition of wireless signals to implement certain families of computational functions on top of the wireless medium. Interestingly, the controllability offered by emerging MS technological solutions has the potential to enable more elaborate computations over-the-air, essentially offloading computations from the communication network's endpoints.
\end{enumerate}

\section{Relevant State-of-the-Art}\label{sec:literature}
The implementation of \gls{DNN}s at the transceivers has been studied under the \gls{JSCC} paradigm, with results that outperform traditional communication systems due to the neural networks' ability to learn useful patterns from the data and channel distributions~\cite{DCH18, BKG19}. However, \gls{JSCC} is limited to data reconstruction as an objective.
Conversely, deep semantic communication approaches \cite{GQA23}, endow the communication system with the purpose of transmitting the meaning of the data, rather than their bit representations, which can be interpreted as a \gls{TOC} objective. 
Notably, the DeepSC architecture of~\cite{DeepSC} proposes a \gls{DNN} splitting approach, with separate source and channel coding sub-modules that are trained independently.
The channel encoder and decoder are trained to maximize an \gls{MI} objective, which, on the one hand, is difficult to evaluate analytically for fading channels, while, on the other hand, maximizing the \gls{MI} additionally implies that the effects of the channel are to be negated instead of being used for computations.
In~\cite{JGM21_Image_Retrieval}, a \gls{TOC} approach with separate source and channel coding modules was developed for image retrieval under \gls{AWGN} and Rayleigh fading, which further exemplified the benefits of separate training of each component rather than \gls{E2E}.
However, this training is possible due to the inserted rate-like part of the loss function that also treats the channel as noise, instead of a computational entity.
Besides, the idea of DNN splitting for \gls{EI} tasks has been investigated in~\cite{dnn-splitting, PBD22_bottleneck} from the information bottleneck viewpoint, to derive optimal network partitioning in uncontrollable wireless channels.

Traditionally, \gls{OAC} methods have been developed to compute aggregate functions over multiple-access channels based on the superpositions of signals~\cite{NG07}, and have been utilized for federated learning tasks, as in~\cite{BXW24}.
Both of these~\gls{OAC} approaches, however, focused on computing a limited, yet useful family of analytical functions that are fundamentally different from the computations that take place in hidden neural network layers.
More relevant to the present work is the methodology of~\cite{DeepOAC}, where the wireless channel is treated as a hidden network layer encompassed by \gls{DNN} layers at the transceivers.
While this \gls{E2E} treatment can be utilized under the context of \gls{TOC}, the fact that the computations imposed by wireless channels are not controllable in the absence of any \gls{MS} flavor, provides limited benefits.
In essence, the proposed \gls{MINN} framework in this paper is general enough to accommodate this setup as a special case, once the \gls{MS}-induced links are ignored in the overall system model. In the numerical evaluation section later on, we compare with this variation to illustrate the benefits of integrating MS(s) as hidden layer(s) for effective mixed digital/analog computation.

The joint consideration of \glspl{MS} and deep learning is a growing body of research. Apart from works that introduce deep learning algorithms to control \glspl{RIS}~\cite{9367575,ASH22,SSA2025} or \gls{SIM}~\cite{HJA24}, cascaded \glspl{MS} have been introduced as \gls{DNN} layers of diffractive implementations in \cite{XYN18, LML23, CHQ24}, towards analog computing hardware that is envisioned to exhibit notable benefits in computational speed and power consumption. In this paper, we are motivated by all-optical neural network implementations, but our framework is differentiated by the fact that the \gls{SIM} layers are assumed to reside within the wireless environment, so that the \gls{E2E} system needs to account for time-varying wireless fading when passing information between the \gls{SIM} and digital network layers at the TX/RX endpoints~\cite{RSR25}. It is noted that purely optical \glspl{DNN} have the limitation that the input data needs to be transferred to the \gls{RF} domain via techniques like holography, illumination, or traditional modulation, which are currently impractical for real-life deployment of such architectures. Finally, in the context of wireless networks, other works have capitalized on sophisticated designs of the responses of the elements of \glspl{MS} to implement wave-domain signal processing~\cite{YCX23_Wave_Computing, OTA25} or multi-access edge computing~\cite{ZYY24} tasks.

Regarding approaches that specifically consider \gls{TOC} or semantic communication problems with the inclusion of \glspl{MS}, a \gls{TOC} approach was recently introduced in~\cite{GJZ24_SIM_TOC}, according to which, the \gls{DNN} layers at the transceivers were implemented via \gls{SIM} layers, in contrast to having the \gls{SIM} as part of the environment, as proposed by this work. Indeed, performing \gls{DNN} computations at the \gls{RF} regime even at the endpoints has the aforementioned benefits, however, the hardware design of such transceivers is far from trivial to be implemented by low-cost \gls{IoT} devices.
Besides, an \gls{MS} placed directly inside the wireless environment may offer more precise control of the propagation medium.
Finally, semantic communications were performed through the assistance of an \gls{RIS} placed at the environment in~\cite{HMC24_Mattias_RIS}.
In that approach, the \gls{RIS} was optimized to maximize an equivalent \gls{SNR} objective, similar to~\cite{DeepSC}, whereas we herein propose to treat the problem in an \gls{E2E} manner.
In the numerical evaluation of Section~\ref{sec:results}, we include a baseline where the \gls{RIS} alongside other components are optimized with respect to the achiavable Shannon rate, and we show that this approach is less effective compared to our \gls{E2E} treatment under the considered system, especially in the low-\gls{SNR} regime.
Despite the related literature of \gls{E2E} architectures for \gls{EI} that potentially take advantage of \gls{MS} capabilities, to the best of our knowledge, the proposed \gls{MINN} framework is the first work to highlight the importance of treating the \gls{MS}-enabled smart wireless channel as a favorable computational machinery embedded inside an \gls{E2E} \gls{DNN} architecture.

\section{System and MS Components Modeling}\label{sec:system-model}
\subsection{System and Received Signal Models}
We consider the uplink of a point-to-point \gls{MIMO} communication system, where a \gls{TX} equipped with $N_t$ transmit antennas wishes to transmit its data to an $N_r$-antenna \gls{RX} on a frame-by-frame basis, where the frames are indexed as $t=1,2,\ldots$ and are possibly unevenly spaced. This communication is enhanced via an \gls{MS} (either an \gls{RIS} or \gls{SIM}), deployed as a standalone node in the wireless environment, whose configuration may change at every discrete time step $t$ upon the command of an abstract controller unit. Without loss of generality, let us assume that the \gls{SIM} is consisted of $M$ thin diffractive layers, each with $N_m$ unit elements, so that it contains $N_{\rm SIM} \triangleq M N_m$ phase-tunable elements in total.
For ease of notation and to present a comprehensive system model, we will additionally use $N_m$ to denote the number of \gls{RIS} tunable elements. Furthermore, let us define the \gls{CFR} matrices at each $t$-th time instance for the \gls{TX}-\gls{RX}, the \gls{TX}-\gls{MS}, and the \gls{MS}-\gls{RX} links as $\mathbf{H}_{\rm D}(t) \in \mathbb{C}^{N_r \times N_t}$, $\mathbf{H}_{\rm 1}(t) \in \mathbb{C}^{N_t \times N_m}$, and $\mathbf{H}_{\rm 2}(t) \in \mathbb{C}^{N_r \times N_m}$,  respectively.
The transmitted signal is expressed as $\myvec{s}(t) \in \mathbb{C}^{N_t \times 1}$, which satisfies a power budget constraint $P\triangleq \mathbb{E}[\| \myvec{s}(t) \|]$.
In fact, we suggest that $\myvec{s}(t)$ represents both the intended, source-coded, and modulated data stream (according to the MIMO spatial multiplexing principle, the number of data symbols need to be $d\leq\min\{N_t,N_r\}$), as well as potential beamforming weights, without making any specific assumptions about the underlying procedures that produced the transmitted signal or the distribution of symbols.

During each $t$-th frame transmission, the \gls{MS} is characterized by its controllable {\em phase configuration} vector
$\myvec{\omega}(t)\in\mathbb{C}^{N_m \times 1}$ in the case of an \gls{RIS} and $\myvec{\omega}(t)\in\mathbb{C}^{N_{\rm SIM} \times 1}$ in the case of \gls{SIM},
while its resulting {\em response configuration} is modeled for the idealized case of unit amplitude as $\boldsymbol{\varphi}(t) \triangleq \exp(-\jmath \myvec{\omega}(t))$.
The effects of the responses of the metamaterials in the cascaded channel are captured via the matrix $\boldsymbol{\Omega}(t) \in \mathbb{C}^{N_m \times N_m}$, the structure of which, will be detailed in the next subsection.
In the remainder of this work, $\boldsymbol{\Omega}(t)$, $\boldsymbol{\varphi}(t)$, and $\myvec{\omega}(t)$ are used as generic notation to describe any of the \gls{RIS} and \gls{SIM} cases, while device-specific notation is introduced wherever needed.
Using the above, the baseband received signal at the \gls{RX} antennas is expressed as follows:
\begin{align}
    \mathbf{y}(t) 
    &\triangleq\left(\mathbf{H}_{\rm D}(t)+\mathbf{H}_{\rm 2}(t)\boldsymbol{\Omega}(t)\mathbf{H}_{\rm 1}^{\dagger} (t)\right)\myvec{s}(t) + \myvec{\tilde{n}}\label{eq:received-signal-RIS_1}\\
    &\triangleq \mathcal{T}(\boldsymbol{\mathcal{H}}(t), \myvec{\varphi}(t), \myvec{s}(t))\label{eq:received-signal-RIS_2},
\end{align}
where $\myvec{\tilde{n}} \in \mathbb{C}^{N_r \times 1}$ denotes the \gls{AWGN} at the \gls{RX}, comprising of \gls{iid} samples drawn from the standard complex normal distribution $\mathcal{CN}(0,\sigma^2)$.
In the sequel, we will be making use of the transmission function $\mathcal{T}(\boldsymbol{\mathcal{H}}(t), \myvec{\varphi}(t), \myvec{s}(t))$ as an abstraction, emphasizing that the wireless medium is treated as a programmable computation. In this definition, we use the notation $\boldsymbol{\mathcal{H}}(t) \triangleq \{\mathbf{H}_{\rm D}(t),  \mathbf{H}_{\rm 1}(t), \mathbf{H}_{\rm 2}(t)\}$ for the instantaneous \gls{CSI}, which is assumed to be readily available to all system nodes. Obviously, this availability implies a recurring channel estimation phase at each $t$-th time step, which may be a challenging prerequisite (see~\cite{ASH22} and references therein). Nevertheless, this assumption allows the focus of this work to be on the training and evaluation of the proposed \gls{MINN} architecture.
Logical future extensions could incorporate the channel estimation phase in the DNN transceiver modules themselves, following \gls{ISAC} principles~\cite{BAL24}. Alternatively, channel-agnostic variations of transceivers will be also proposed and evaluated in the following sections to illustrate the performance trade-offs when integrating \glspl{MS} as over-the-air neural network layers.

\subsection{RIS and SIM Models}
Considering first an \gls{RIS}, let its phase configuration vector at time $t$ be denoted as
$\myvec{\theta}(t) \triangleq [\theta_1(t),\theta_2(t),\ldots,\theta_{N_m}(t)]^\top$
($\equiv\myvec{\omega}(t)$ in \eqref{eq:received-signal-RIS_2}), so that the phase state of its $n$-th unit element ($n=1,2,\ldots,N_m$) is expressed as $\theta_n(t) \in [0, 2\pi)$.
Then, the induced weights at the response configuration vector are given as $\myvec{\phi}(t) \triangleq \exp(-\jmath \theta(t))$ ($\equiv\myvec{\varphi}(t)$)
In this case, using the diagonal matrix definition $\myvec{\Phi}(t) \triangleq {\rm diag}(\myvec{\phi}(t))\in \mathbb{C}^{N_m \times N_m}$, it holds that $\myvec{\Omega}(t)\equiv \myvec{\Phi}(t)$ in~\eqref{eq:received-signal-RIS_1}. 

Proceeding to the introduction of the \gls{SIM} into the system model, we first assume that all $M$ layers ($m=1,2,\ldots,M$) are closely stacked and aligned parallel to each other, with their shared normal vector oriented perpendicular to the line connecting the \gls{TX} and \gls{RX} positions. Under this placement, the signal from the \gls{TX} arrives at the first layer of the \gls{SIM}, undergoes diffraction and controllable phase shifting by the consecutive $M-1$ layers, before being finally diffracted towards the \gls{RX}. Let us define the distance between any consecutive \gls{MS} layers as $d_M$ and the area of each unit element as $S_M$.
Due to the compact placement of the layers, the layer-to-layer propagation can be accurately modeled via the Rayleigh-Sommerfeld diffraction equation \cite{GJZ24_SIM_TOC, AXN23}.
Namely, we define the propagation coefficient matrix from each $(m-1)$-th to the $m$-th layer as $\myvec{\Xi}_m \in \mathbb{C}^{N_m \times N_m}$, so that its $(n, n')$-th entry ($n,n'=1,2,
\ldots,N_m$) includes the propagation gain between the (arbitrarily ordered) $n$-th unit element of the $(m-1)$-th layer and the $n'$-th element of the next layer, as follows~\cite{AXN23}:
\begin{align}
    [\myvec{\Xi}_m]_{n,n'} &\triangleq \frac{d_M S_M}{(d_{n,n'})^2} 
    \Big( \frac{1}{2\pi d_{n,n'}} - \frac{\jmath}{\lambda} \Big) 
    \exp({\jmath 2\pi d_{n,n'}}),
\end{align}
where $ d_{n,n'}$ denotes the distance between the centers of the $n$-th and $n'$-th elements, and $\lambda$ is the carrier frequency.

Apart from diffracting, each $n$-th element of each $m$-th \gls{SIM} layer introduces a controllable weight, similar to the \gls{RIS} modeling, denoted as $[\myvec{\psi}_m(t)]_n \triangleq \exp(-\jmath \vartheta^m_n(t))$. 
We also introduce $\myvec{\psi}_m(t)$ including the {\em response configuration} at the $m$-th layer, the overall response configuration $\myvec{\psi}(t) = [\myvec{\psi}^\top_1(t),\myvec{\psi}^\top_2(t),\dots,\myvec{\psi}^\top_M(t)]^\top \in \mathbb{C}^{N_{\rm SIM} \times 1}$, and the {\em phase configuration} vector of the \gls{SIM} $\myvec{\vartheta}(t) \triangleq [\vartheta^1_1(t), \dots, \vartheta^1_{N_m}(t), \dots, \vartheta^M_1(t), \dots, \vartheta^M_{N_m}(t)]^\top \in \mathbb{C}^{N_{\rm SIM} \times 1}$. By defining $\myvec{\Psi}_m(t) \triangleq {\rm diag}(\myvec{\psi}_m(t))$, the overall \gls{SIM} response can be mathematically expressed via the following matrix~\cite{RSR25}:
\begin{equation}
 \myvec{\Upsilon}(t) \triangleq\left(\prod_{m=M}^{2} \myvec{\Psi}_m(t) \myvec{\Xi}_m \right) \myvec{\Psi}_1(t)\in\mathbb{C}^{N_m \times N_m}.
\end{equation}
Note that, for $\myvec{\Omega}(t)\equiv \myvec{\Upsilon}(t)$, \eqref{eq:received-signal-RIS_1} holds for the \gls{SIM} case. 

Revisiting the previously introduced generic notations of $\myvec{\Omega}(t)$ and $\myvec{\omega}(t)$, they can now be expressed concretely for each of the \gls{RIS} and \gls{SIM} cases as $\myvec{\Omega}(t) \in \{ \myvec{\Phi}(t), \myvec{\Upsilon}(t) \}$ and $\myvec{\omega}(t) \in \{ \myvec{\theta}(t), \myvec{\vartheta}(t) \}$.
In the rest of this paper, $\myvec{\Omega}(t)$, $\myvec{\omega}(t)$, and $\myvec{\varphi}(t)$ will be used when the underlying operations are agnostic of the \gls{MS} type, while $\myvec{\Phi}(t)$, $\myvec{\Upsilon}(t)$, and their respective vectors will be explicitly utilized when the operations need to discriminate between \glspl{RIS} and \gls{SIM}.

\section{Metasurface-Integrated Neural Networks}\label{sec:architecture}
\begin{figure*}[t]
\centering
    \includegraphics[width=0.85\textwidth]{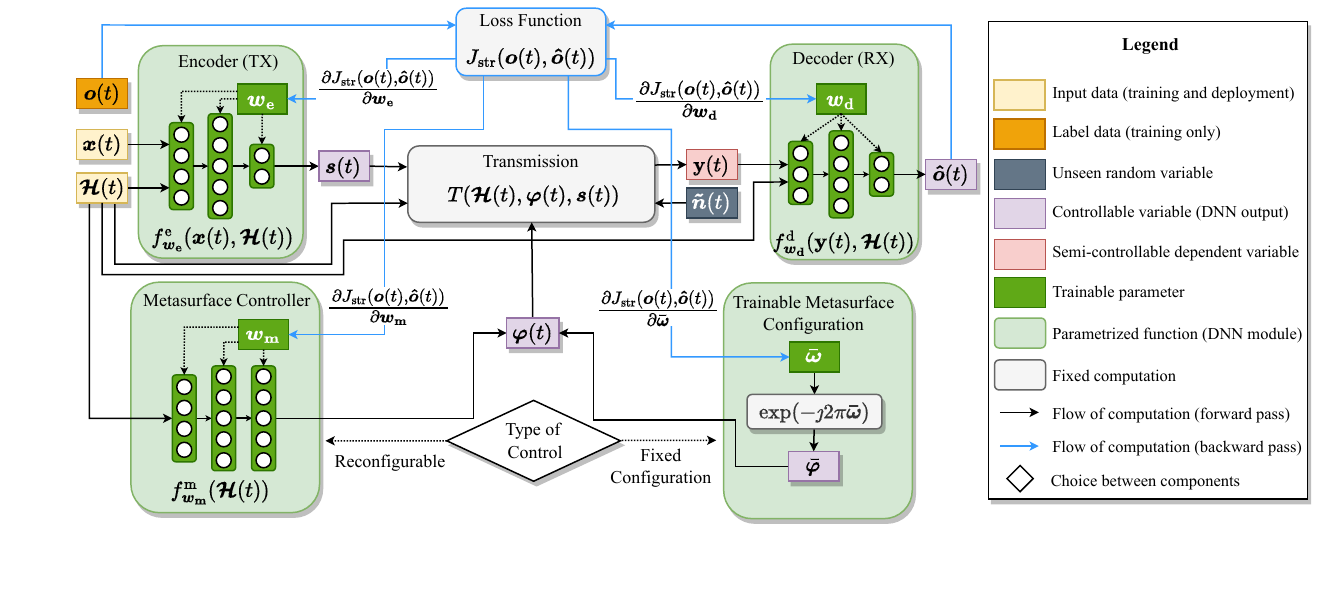}
    \vspace{-0.5cm}
    \caption{Block diagram and computation flow for the proposed \gls{E2E} \gls{TOC} framework where the metasurface-parametrizable channel acts as an intermediate \gls{DNN} component. Both the cases of reconfigurable and static metasurfaces are included, entailing different procedures during the forward and backward passes.}
    \label{fig:architecture}
\end{figure*}

\subsection{Transceiver Modules}
As initially discussed in Section~\ref{sec:EI-theory}, \gls{EI} entails two computational modules, collocated at the transceiver endpoints.
To implement the {\em ``infer-while-transmitting''} methodology, the \gls{TX} utilizes an encoder \gls{DNN}, $f^{\rm e}_{\myvec{w_{\rm e}}}(\cdot)$, providing the output $\myvec{s}(t)$, while the \gls{RX} operates a decoder \gls{DNN}, $f^{\rm d}_{\myvec{w_{\rm d}}}(\cdot)$, deriving the output $\myvec{\hat{o}}(t)$.
Those blocks are tasked with performing compression, encoding and decoding, error resilience and correction, and potential transmit and receive beamforming alongside probabilistic inference. The exact layer architecture of those models is purposely left unspecified at this stage as a practitioner's choice, depending, in general, on: \textit{i}) the nature of the wireless environment; \textit{ii}) the type of input and target data; \textit{iii}) computational capabilities of the transceivers' hardware; as well as \textit{iv}) the current state-of-the-art. We only note that different sub-modules may be used for each of the above operations, while, typically for uplink scenarios, $f^{\rm d}_{\myvec{w_{\rm d}}}(\cdot)$ can be implemented with larger \gls{DNN} structures due to the constant power supply at base stations. In addition, regardless of the choice of neural network, we impose a final fixed post-processing step at the encoder's output $\myvec{s}(t)$ to satisfy the TX system's power budget, as follows:
\begin{equation}\label{eq:power-norm}
    \myvec{s}(t) \gets \sqrt{P}\frac{\myvec{s}(t)}{\|\myvec{s}(t)\|}. 
\end{equation}

Considering the concrete input arguments of the encoder functions, two different variations may be defined depending on whether \gls{CSI} is available to each of the endpoints.

\subsubsection{Channel-Agnostic Transceivers}
An instance of the data variables $\myvec{x}(t)$ is observed at the \gls{TX}, that is passed to the encoder to construct the transmitted signal, while the decoder \gls{DNN} observes the received signal and performs an estimate of the unseen target variable $\myvec{o}(t)$; this can be described as:
\begin{align}
    \myvec{s}(t) &= f^{\rm e}_{\myvec{w_{\rm e}}}(\myvec{x}(t))\label{eq:encoder-channel-agnostc}, \\
    \myvec{\hat{o}}(t) &= f^{\rm d}_{\myvec{w_{\rm d}}}(\mathbf{y}(t)) \label{eq:decoder-channel-agnostc}.
\end{align}
Since no CSI is used by the endpoints, we highlight the similarity of this design to source-only coding, despite the fact that the encoder may need to add redundancy in the transmitted signal, which is traditionally considered as channel coding.
Evidently, both processes need to guarantee that the inference procedure performs sufficiently, irrespective of the current channel conditions, which may be a demanding request.
Nevertheless, not requiring \gls{CSI} is a strong simplification of the system architecture, therefore, it is included later on in our investigations in this paper.

\subsubsection{Channel-Aware Transceivers}
Let us assume a quasi-static fading channel and a channel estimation procedure that takes place within each $t$-th channel frame before data transmission, based on which the \gls{TX} and \gls{RX} modules obtain accurate estimates of the \gls{CFR} matrices $\boldsymbol{\mathcal{H}}(t)$. Each module may receive $\boldsymbol{\mathcal{H}}(t)$ as an additional input, yielding respectively the following representations for the encoder/decoder DNNs:
\begin{align}
    \myvec{s}(t) &= f^{\rm e}_{\myvec{w_{\rm e}}}(\myvec{x}(t), \boldsymbol{\mathcal{H}}(t))\label{eq:encoder-channel-aware}, \\
    \myvec{\hat{o}}(t) &= f^{\rm d}_{\myvec{w_{\rm d}}}(\mathbf{y}(t), \boldsymbol{\mathcal{H}}(t)) \label{eq:decoder-channel-aware}.
\end{align}

Endowing the TX/RX modules with \gls{CSI} leads to more resilient transmission schemes that closely resemble \gls{JSCC} \cite{GWT24_JSCC_Review}.
The main difference lies in that \gls{JSCC} focuses on estimating $\myvec{x}(t)$, while \gls{EI} deals with approximating $\myvec{o}(t) = l(\myvec{x}(t))$\footnote{Those two problems can be considered equivalent by setting the mapping function $l(\cdot)$ to be the identity function, yielding $\myvec{o}(t) = \myvec{x}(t)$, and adopting \gls{MSE} as the objective function $J_{\rm str}(\cdot)$. As a result, \gls{EI} is a more general problem formulation than communications which focus on data reconstruction.}.
In this paper, we assume that channel estimation takes place transparently before every transmission, both during training and inference ,and results in noise-free estimations of $\boldsymbol{\mathcal{H}}(t)$.
Accounting for noisy estimates or even incorporate the estimation in the procedure under \gls{ISAC} paradigms lead exciting research directions, which we will study in future works.

\subsection{Control Module for Reconfigurable Metasurfaces}
When \gls{CSI} is available, as in most wireless communication settings, the \gls{MS} changes its response configuration at every transmission frame to optimize the system's objective~\cite{10600711}.
To incorporate this mode of operation into our \gls{E2E} architecture, $\myvec{\varphi}(t)$ is treated as a controllable variable that is the output of a third \gls{DNN} module.
Specifically, we define the {\em metasurface controller} as the following neural network:
\begin{equation} \label{eq:ris-controller}
    \myvec{\varphi}(t) = f^{\rm m}_{\myvec{w_{\rm m}}}(\boldsymbol{\mathcal{H}}(t)),
\end{equation}
imposing that the final layer performs the operation $\myvec{\varphi}(t) = \exp(-\jmath \myvec{\hat{\vartheta}})$. 
As stated before, either $\myvec{\phi}(t)$ or $\myvec{\psi}(t)$ may be the actual output of the module depending of the selected type of \gls{MS}, however, we keep the abstract notation of $\myvec{\varphi}(t)$ to provide a general framework.
Under this viewpoint, the \gls{MS} is a controllable entity that can be adapted dynamically to offer favorable wave-domain computation at every channel realization.
This treatment allows for fine-grained control over the reprogrammability of the environment, at the cost of an additional neural network module and the associated hardware requirements.
Plugging the three trained modules from expressions~\eqref{eq:encoder-channel-agnostc}--\eqref{eq:ris-controller} 
onto the received signal in~\eqref{eq:received-signal-RIS_1}, we can derive the \gls{E2E} inference model $\myvec{\hat{o}}(t)=f^{\rm r}_{\myvec{w_{\rm r}}}(\myvec{x}(t), \boldsymbol{\mathcal{H}}(t))$, for the two channel knowledge cases, as follows:
\begin{equation}\label{eq:e2e-model-reconfigurable}
\footnotesize 
\myvec{\hat{o}}(t)\!=\!\begin{cases}
    \underbrace{f^{\rm d}_{\myvec{w_{\rm d}}} \bigg(\mathcal{T}\big(\boldsymbol{\mathcal{H}}(t), f^{\rm m}_{\myvec{w_{\rm m}}}(\boldsymbol{\mathcal{H}}), f^{\rm e}_{\myvec{w_{\rm e}}}(\myvec{x}(t))\big)\bigg),}_{\text{channel-agnostic transceivers}} \\
    \underbrace{f^{\rm d}_{\myvec{w_{\rm d}}} \Big(\mathcal{T}\big(\boldsymbol{\mathcal{H}}(t), f^{\rm m}_{\myvec{w_{\rm m}}}(\boldsymbol{\mathcal{H}}(t)), f^{\rm e}_{\myvec{w_{\rm e}}}(\myvec{x}(t),\boldsymbol{\mathcal{H}}(t))\big), \boldsymbol{\mathcal{H}}(t)\Big),}_{\text{channel-aware transceivers}} 
\end{cases}    
\end{equation}
where the overall trainable weights of this reconfigurable architecture have been represented as $\myvec{w_{\rm r}} \triangleq \{\myvec{w_{\rm d}}, \myvec{w_{\rm e}}, \myvec{w_{\rm m}}\}$, which can be trained together under the same objective functions and backward passes, as it will be detailed in Section~\ref{sec:training-deployment}.

\subsection{Metasurfaces with Trainable Fixed Response}
As an alternative approach, one may choose to directly learn a fixed configuration for the \gls{MS}; let us denote this as $\myvec{\bar{\omega}}$.
While the training process may iteratively evaluate multiple candidate values for $\myvec{\bar{\omega}}$, once the training is complete, the learned configuration is equipped onto the \gls{MS} to maintain a constant (static) response configuration $\myvec{\varphi}(t) \equiv \myvec{\bar{\varphi}} \triangleq \exp(-\jmath \myvec{\bar{\omega}})$ over time, irrespective of the channel conditions or input data.
This description is more akin to the idea that the effective phase configurations are treated similarly to \gls{DNN} weights, as they too remain fixed after the completion of the training procedure, and are used to perform the same computational operations over varying input instances. To this end, denote the fixed phase configurations of the \gls{RIS} and \gls{SIM} as $\myvec{\bar{\theta}}$ and $\myvec{\bar{\vartheta}}$, respectively.
The training procedure optimizes $\myvec{\bar{\omega}}$ directly, i.e., its weights $\myvec{w_{\rm s}} \triangleq \{\myvec{w_{\rm d}}, \myvec{w_{\rm e}}, \myvec{\bar{\omega}}\}$, therefore, the \gls{E2E} static architecture can be now expressed as follows:
\begin{equation}\label{eq:e2e-model-static}
\footnotesize
    \myvec{\hat{o}}(t) = 
    \begin{cases}
        \underbrace{f^{\rm d}_{\myvec{w_{\rm d}}} \left(\mathcal{T}\left(\boldsymbol{\mathcal{H}}(t), \myvec{\bar{\varphi}}, f^{\rm e}_{\myvec{w_{\rm e}}}(\myvec{x}(t))\right)\right),}_{\text{channel-agnostic transceivers}} \\
        \underbrace{f^{\rm d}_{\myvec{w_{\rm d}}} \bigg(\mathcal{T}\big(\boldsymbol{\mathcal{H}}(t), \myvec{\bar{\varphi}}, f^{\rm e}_{\myvec{w_{\rm e}}}(\myvec{x}(t),\boldsymbol{\mathcal{H}}(t))\big), \boldsymbol{\mathcal{H}}(t)\bigg).}_{\text{channel-aware transceivers}}
    \end{cases} 
\end{equation}

It is noted that, while reconfigurable \glspl{MS} may offer more precise control in shaping the exact form of $\mathcal{T}(\cdot)$, the extra hidden layers required by the inclusion of the \gls{MS} controller module may well hinder the training capabilities of the proposed \gls{MINN} compared to the current variation.
Besides, assuming wireless systems of reasonably limited variability, such as \gls{LoS} dominant environments with fixed transceivers, static \gls{MS} configurations may offer satisfactory performance.
The next section addresses the systemic requirements for all variations of this section, while performance trade-offs are investigated under our numerical evaluations.

The final thing to consider about the proposed \gls{MINN} framework are the theoretical approximation guarantees, especially considering its universal approximation capabilities.
Presenting a concrete analysis of this property lies well outside the scope of this paper, as it involves proving that functionals derived of the modeling of Section~\ref{sec:system-model} are discriminatory and dense in the space of continuous complex functions as in~\cite{Cybenko89}, while also accounting for the stochastic nature of both the fading components and the \gls{AWGN}.
Nevertheless, we argue that, since our \gls{MINN} architecture includes two typical \glspl{DNN} at the start and the end of the cascaded computations, which are in fact universal approximators, the \gls{E2E} architecture should also contain this property at least in the infinite-\gls{SNR} regime. Intuitively, as long as $\mathcal{T}(\cdot)$ does not lose information due to the noisy channel, the decoder \gls{DNN} should be capable of approximating the optimal decoder of $\mathbf{y}(t)$ to obtain $\boldsymbol{s}(t)$, so that the channel is completely negated, and the \gls{E2E} \gls{MINN} reduces to $f^{\rm d}_{\myvec{w_{\rm d}}} \big(f^{\rm e}_{\myvec{w_{\rm e}}}(\myvec{x}(t),\boldsymbol{\mathcal{H}}(t)), \boldsymbol{\mathcal{H}}(t)\big)$, which is indeed a universal approximator consisting of standard \gls{DNN} layers.

\section{MINN Training and Deployment}\label{sec:training-deployment}
To perform neural network training at any of the wireless communication system nodes, a separate data collection step is carried out to generate a set of $|\mathcal{D}|$ labeled data instances: $\mathcal{D} \triangleq \{ (\myvec{x}_i, \myvec{o}_i) \}_{i=1}^{|\mathcal{D}|}$. Let us also assume the availability of a set of $|\mathcal{C}|$ channel sample estimates (in respective coherent time instances): $\mathcal{C} \triangleq \{\boldsymbol{\mathcal{H}}(t) \}_{t=1}^{|\mathcal{C}|}$, not necessarily equally spaced. In this paper, we make the assumption that the channel realizations are conditionally independent\footnote{In certain scenarios, the data realizations and the statistical properties of the channel can be statistically dependent. For example, in a target detection system where the observations $\myvec{x}_i$ contain sensory inputs, while $\myvec{o}_i$ is a binary variable indicating the existence of a target in the area of interest, deep fading may be encountered more often when a target is subject to signal blockages. In such cases, the two collection processes of channel measurements and observed data must be synchronous, and a more detailed formulation of the \gls{EI} objective is required. However, the inference problem itself may be potentially computationally easier, since the \gls{CSI} observation provides additional information regarding the target value.} from $\mathcal{D}$'s data instances. It is noted that, while this is a rather lenient assumption, it is crucial in permitting the evaluation of the expectation in $\mathcal{OP}_{\rm EI}$'s objective via \gls{iid} Monte Carlo samples.  

\subsection{Backpropagation Over the Wireless Channel}\label{subsec:Training}
The training procedure can be described as a variation of the standard gradient descent approach for neural network training, with the inclusion of channel samples.
To provide a comprehensive framework, let us use the generic parameter vector $\myvec{w_{\rm k}}$ with $\mathbf{k} \in \{\mathbf{r},\mathbf{s}\}$, taking the form of either $\myvec{w_{\rm r}}$ or $\myvec{w_{\rm s}}$ depending on the choice of use of reconfigurable or static \glspl{MS}.
Similar to standard deep learning practices, our \gls{E2E} \gls{MINN} architecture may be optimized using \gls{SGD} over the collected data and channel instances.
Specifically, let us express the data-channel as $J_{\rm str}(\myvec{o}_i, \myvec{\hat{o}}_i) = J_{\rm str}(\myvec{o}_i, f^{\rm k}_{\myvec{w_{\rm k}}}(\myvec{x}(t), \boldsymbol{\mathcal{H}}(t)))$ to explicit show the dependence of the loss function on the instantaneous wireless channel conditions. To this end, leveraging the previous described conditional independence assumption, $\mathcal{OP}_{\rm EI}$'s objective can be approximated as follows:
\begin{equation}\label{eq:SGD-objective}
    \mathbb{E}_{\boldsymbol{\mathcal{H}}}[J(\myvec{w_{\rm k}})]\!\cong\!\frac{1}{|\mathcal{C}||\mathcal{D}|} \sum_{t=1}^{|\mathcal{C} |}   \sum_{i=1}^{|\mathcal{D}|} J_{\rm str}(\myvec{o}_i, f^{\rm k}_{\myvec{w_{\rm k}}}(\myvec{x}_i, \boldsymbol{\mathcal{H}}(t))).
\end{equation}
In the online version of \gls{SGD}, at every time $t$, one may select a single data point and channel instance to evaluate \eqref{eq:SGD-objective}, and accordingly update the parameter vector as follows:
\begin{equation}\label{eq:w-update}
    \myvec{w_{\rm k}} \gets \myvec{w_{\rm k}} - \eta \nabla_{\myvec{w_{\rm k}}}J_{\rm str}(\myvec{o}(t), f^{\rm k}_{\myvec{w_{\rm k}}}(\myvec{x}(t), \boldsymbol{\mathcal{H}}(t))),
\end{equation}
for some chosen learning rate $\eta$, with the gradient at each case being defined by one of the two following expressions:
\begin{align}
    \nabla J_{\rm str} &= \underbrace{\bigg[ \Big[\frac{\partial J_{\rm str}}{\partial \myvec{w_{\rm d}}}\Big]^\top, \Big[\frac{\partial J_{\rm str}}{\partial \myvec{w_{\rm e}}}\Big]^\top, \Big[\frac{\partial J_{\rm str}}{\partial \myvec{w_{\rm m}}}\Big]^\top\bigg]^\top}_{\text{reconfigurable metasurface}}
    \\
    \nabla J_{\rm str} &= \underbrace{\bigg[ \Big[\frac{\partial J_{\rm str}}{\partial \myvec{w_{\rm d}}}\Big]^\top, \Big[\frac{\partial J_{\rm str}}{\partial \myvec{w_{\rm e}}}\Big]^\top, \Big[\frac{\partial J_{\rm str}}{\partial \myvec{\bar{\omega}}}\Big]^\top\bigg]^\top}_{\text{metasurface with trainable fixed response}}.
\end{align}

Under the \gls{iid} sampling assumption, the consecutive evaluations of the gradient of the objective function in~\eqref{eq:w-update} at each training instance $t$ are unbiased estimators of the true gradient of the objective in~\eqref{eq:SGD-objective}.
Therefore, following the stochastic approximation framework~\cite{Robbins_Monro}, the repetition of this procedure will converge to the true value of the expectation with probability $1$ up to a precision of $O(\eta)$ around it, using constant step size~\cite{Borkar2008}. 
The complete training procedure is detailed in Algorithm~\ref{alg:SGD}, which supports all variations of channel-agnostic/-aware transceivers, static/reconfigurable \gls{MS} controllers, and \gls{RIS}/\gls{SIM} structure.
Lines $6$-$9$ implement our \gls{MINN} architecture as defined in \eqref{eq:e2e-model-reconfigurable}~and~\eqref{eq:e2e-model-static}.
Naturally, batched gradient descent versions may be used alongside more elaborate gradient updates, such as momentum, weight decay (regularization), and adaptive rates \cite{Adam}, however, such implementation details have been left out for ease of presentation.

\begin{algorithm}[t]
\caption{Training of the Proposed \gls{E2E} \gls{MINN}}
\label{alg:SGD}
\begin{algorithmic}[1]
\State Construct DNN weight vector $\myvec{w}$ as one of the following:
\Statex \hspace{1em} \textit{i}) $\myvec{w_{\rm k}} = {\rm concat}(\myvec{w_{\rm d}}, \myvec{w_{\rm e}}, \myvec{w_{\rm m}})$. \hspace{1em} \Comment{$\myvec{w_{\rm k}} \gets \myvec{w_{\rm r}}$}
\Statex \hspace{1em} \textit{ii}) $\myvec{w_{\rm k}} = {\rm concat}(\myvec{w_{\rm d}}, \myvec{w_{\rm e}}, \myvec{\bar{\omega}})$. \hspace{1em} \Comment{$\myvec{w_{\rm k}} \gets \myvec{w_{\rm s}}$}
\State Initialize $\myvec{w}$ randomly.
\For{$t = 1, 2, \ldots, $ until convergence}
    \State Sample $(\myvec{x}(t), \myvec{o}(t))$ from $\mathcal{D}$.
    \State Sample $\boldsymbol{\mathcal{H}}(t)$ from $\mathcal{C}$.
    \State Compute $\myvec{s}(t)$ using one of the following:
    \Statex \hspace{2em} \textit{i}) $\myvec{s}(t) = f^{\rm e}_{\myvec{w_{\rm e}}}(\myvec{x}(t))$. \hspace{2em} \Comment{eq.~\eqref{eq:encoder-channel-agnostc}}
    \Statex \hspace{2em} \textit{ii}) $\myvec{s}(t) = f^{\rm e}_{\myvec{w_{\rm e}}}(\myvec{x}(t), \boldsymbol{\mathcal{H}}(t))$. \hspace{2em} \Comment{eq.~\eqref{eq:encoder-channel-aware}}
    \State Compute $\myvec{\phi}(t)$ using one of the following:
    \Statex \hspace{2em} \textit{i}) $\myvec{\varphi}(t) = f^{\rm m}_{\myvec{w_{\rm m}}}(\boldsymbol{\mathcal{H}}(t))$. \hspace{2em} \Comment{eq.~\eqref{eq:ris-controller}}
    \Statex \hspace{2em} \textit{ii}) $\myvec{\varphi}(t) = \boldsymbol{\bar{\varphi}}$.
    \State Transmit $\myvec{s}(t)$ to receive $\mathbf{y}(t)$: 
    \Statex \hspace{2em} $\mathbf{y}(t) = \mathcal{T}(\boldsymbol{\mathcal{H}}(t), \myvec{\varphi}(t), \myvec{s}(t))$. \Comment{eq.~\eqref{eq:received-signal-RIS_2}}
    \State Compute $\myvec{\hat{o}}(t)$ using one of the following:
    \Statex \hspace{2em} \textit{i}) $\myvec{\hat{o}}(t) = f^{\rm d}_{\myvec{w_{\rm d}}}(\mathbf{y}(t))$. \hspace{2em} \Comment{eq.~\eqref{eq:decoder-channel-agnostc}}
    \Statex \hspace{2em} \textit{ii}) $\myvec{\hat{o}}(t) = f^{\rm d}_{\myvec{w_{\rm d}}}(\mathbf{y}(t), \boldsymbol{\mathcal{H}}(t))$. \hspace{2em} \Comment{eq.~\eqref{eq:decoder-channel-aware}}
    \State Set $\myvec{w_{\rm k}}\!\gets\!\myvec{w_{\rm k}}\!-\!\eta \nabla_{\myvec{w_{\rm k}}}J_{\rm str}(\myvec{o}(t), f^{\rm k}_{\myvec{w_{\rm k}}}(\myvec{x}(t), \boldsymbol{\mathcal{H}}(t)))$. 
\EndFor
\State \Return $\myvec{w}$
\end{algorithmic}
\end{algorithm}

The crux of the training procedure is the gradient update mechanism of \eqref{eq:w-update}.
Since \eqref{eq:e2e-model-reconfigurable} and~\eqref{eq:e2e-model-static} are differentiable operations with respect to $\myvec{w_{\rm s}}$ or $\myvec{w_{\rm r}}$, the partial derivatives may be computed via automatic differentiation tools, by applying the chain rule on the underlying computational graph. Regardless, for the shake of completeness, we provide the derivations for the partial derivatives of the various modules, however, treating the implementation-defined derivatives of the classical neural network components (i.e., $\partial f_{\myvec{w}_{\mathrm{e}}}^{\mathrm{e}}/\partial \myvec{w}_{\mathrm{e}}$, $\partial f_{\myvec{w}_{\mathrm{d}}}^{\mathrm{d}}/\partial \myvec{w}_{\mathrm{d}}$, $\partial f_{\myvec{w}_{\mathrm{m}}}^{\mathrm{m}}/\partial \myvec{w}_{\mathrm{m}}$, and $\partial f_{\myvec{w}_{\mathrm{d}}}^{\mathrm{d}} / \partial \mathbf{y}(t)$) as known.
Continuing, we will make use of the identity $\textrm{vec}(\mathbf{A} \mathbf{X} \mathbf{B}) = (\mathbf{B}^\top \otimes \mathbf{A}) \textrm{vec}(\mathbf{X})$ and that, for an $n$-element vector $\myvec{x}$ and $\myvec{X} = {\rm diag}(\myvec{x})$, the vectorization operation on $\myvec{X}$ can be expressed using matrix operations as ${\rm vec}(\myvec{X}) = \myvec{D}\myvec{x}$, where $\myvec{D} \triangleq [\myvec{D}_1,\myvec{D}_2,\ldots,\myvec{D}_n]$ is an $n^2 \times n$ matrix used for selecting the diagonal elements, in which $\myvec{D}_i$ is an $n \times n$ matrix with binary elements having $1$ at its $(i,i)$-th element and $0$ elsewhere. 

For the case of a reconfigurable \gls{MS} (either an \gls{RIS} or \gls{SIM}), $\myvec{\hat{o}}$ is computed via  \eqref{eq:e2e-model-reconfigurable}.
By applying backwards propagation, the following derivations are deduced:
\begin{align}
    \frac{\partial J_{\rm str}}{\partial \myvec{w_{\rm d}}} 
    &= \frac{\partial J_{\rm str}}{\partial \hat{\myvec{o}}(t)} \cdot \frac{\partial f_{\myvec{w_{\rm d}}}^{\rm d}}{\partial \myvec{w_{\rm d}}}, \label{eq:backprop_d} \\ 
    \frac{\partial J_{\rm str}}{\partial \myvec{w_{\rm m}}} 
    &= \frac{\partial J_{\rm str}}{\partial \hat{\myvec{o}}(t)} \cdot
    \frac{\partial f_{\myvec{w_{\rm d}}}^{\rm d}}{\partial \mathbf{y}(t)}\cdot
    \frac{\partial  \mathbf{y}(t)}{\partial f^{\rm m}_{\myvec{w_{\rm m}}}}\cdot
    \frac{\partial  f^{\rm m}_{\myvec{w_{\rm m}}}}{\partial \myvec{w_{\rm m}}}, \label{eq:backprop_m} \\
    \frac{\partial J_{\rm str}}{\partial \myvec{w_{\rm e}}} 
    &= \frac{\partial J_{\rm str}}{\partial \hat{\myvec{o}}(t)} \cdot
    \frac{\partial f_{\myvec{w_{\rm d}}}^{\rm d}}{\partial \mathbf{y}(t)} \cdot
    \frac{\partial \mathbf{y}(t)}{\partial f_{\myvec{w_{\rm e}}}^{\rm e}} \cdot
    \frac{\partial f_{\myvec{w_{\rm e}}}^{\rm e}}{\partial \myvec{w_{\rm e}}} \label{eq:backprop_e},
\end{align}
where $\partial J_{\rm str}/\partial \hat{\myvec{o}}(t)$ concerns the gradient of the problem-defined loss function with respect to the network's output, which, for the example of the \gls{CE} loss of~\eqref{eq:cross-entropy}, is computed as $-\myvec{o}(t)/\myvec{\hat{o}}(t)$.
The remaining terms are defined as follows:
\begin{align}
    \frac{\partial \mathbf{y}(t)}{\partial f_{\myvec{w_{\rm e}}}^{\rm e}} &= \mathbf{H}_{\rm 2}(t)\boldsymbol{\Phi}(t)\mathbf{H}_{\rm 1}^{\dagger} (t) + \mathbf{H}_{\rm D}(t), \\
    \frac{\partial \mathbf{y}(t)}{\partial f^{\rm m}_{\myvec{w_{\rm m}}}} &=\frac{\partial \mathbf{y}(t)}{\partial \myvec{\varphi}(t)} = \big((\myvec{s}^{\top}(t)\mathbf{H}_{\rm 1}^{\ast}(t)) \otimes \mathbf{H}_{\rm 2}(t)\big)\myvec{D} \label{eq:phi-gradient},
\end{align}
with $\myvec{D}$ being the $N_m^2 \times N_m$ binary selection matrix. 

In the fixed-configuration \gls{RIS} case, $\myvec{\hat{o}}$ is computed via~\eqref{eq:e2e-model-static}, and the trainable configuration may be concretely expressed as $\myvec{\bar{\omega}} \equiv \myvec{\bar{\theta}}$; the phase shift vector is $\myvec{\bar{\varphi}} \equiv \myvec{\bar{\phi}}$.
The quantities $\partial J_{\rm str}/\partial \myvec{w_{\rm e}}$ and $\partial J_{\rm str}/\partial \myvec{w_{\rm e}}$ remain the same, yielding:
\begin{align}
    \frac{\partial J_{\rm str}}{\partial \myvec{\bar{\omega}}}
    = \frac{\partial J_{\rm str}}{\partial \myvec{\bar{\theta}}}
    =  \frac{\partial J_{\rm str}}{\partial \hat{\myvec{o}}(t)} \cdot
    \frac{\partial f_{\myvec{w_{\rm d}}}^{\mathrm{d}}}{\partial \mathbf{y}(t)} \cdot
    \frac{\partial \mathbf{y}(t)}{\partial \myvec{\bar{\phi}}} \cdot
    \frac{\partial \myvec{\bar{\phi}}}{\partial \myvec{\bar{\theta}}},
    \label{eq:backprop_ris}
\end{align}
where $\partial \mathbf{y}(t)/\partial \myvec{\bar{\phi}}$ can be computed as in~\eqref{eq:phi-gradient}, while
$\partial\myvec{\bar{\phi}} / \partial \myvec{\bar{\theta}} = -\jmath \exp{(-\jmath \myvec{\bar{\theta}})}$.

For the fixed-configuration \gls{SIM} case, the trainable configuration is expressed as $\myvec{\bar{\omega}} \equiv\myvec{\bar{\vartheta}}$, while the static phase shifts are $\myvec{\bar{\varphi}} \equiv \myvec{\bar{\psi}}$, and again, $\myvec{\hat{o}}$ is computed via~\eqref{eq:e2e-model-static}.
Similarly, the following derivations hold:
\begin{align}
    \frac{\partial J_{\rm str}}{\partial \myvec{\bar{\omega}}}
    = \frac{\partial J_{\rm str}}{\partial \myvec{\bar{\vartheta}}}
    =  \frac{\partial J_{\rm str}}{\partial \hat{\myvec{o}}(t)} \cdot
    \frac{\partial f_{\myvec{w_{\rm d}}}^{\mathrm{d}}}{\partial \mathbf{y}(t)} \cdot
    \frac{\partial \mathbf{y}(t)}{\partial  \myvec{\bar{\psi}}} \cdot
    \frac{\partial  \myvec{\bar{\psi}}}{\partial \myvec{\bar{\vartheta}}},
    \label{eq:backprop_sim}
\end{align}
where $ \partial J_{\rm str} / \partial \hat{\myvec{o}}(t)$ and $\partial f_{\myvec{w}_{\mathrm{d}}}^{\mathrm{d}} / \partial \mathbf{y}(t)$ are the same as before, while, similarly, $\partial  \myvec{\bar{\psi}}/\partial \myvec{\bar{\vartheta}} = -\jmath \exp{(-\jmath \myvec{\bar{\vartheta}})}$.
Since now $\mathcal{T}(\cdot)$ involves the \gls{SIM} system model, $\partial \mathbf{y}(t) / \partial  \myvec{\bar{\psi}}$ requires further derivations.
Following the same procedure as in~\eqref{eq:phi-gradient}, and by denoting the response matrix of each of the $M$ \gls{SIM} elements as $\myvec{\bar{\Psi}}_m \triangleq {\rm diag}( \myvec{\bar{\psi}}_m)$, it is deduced $\partial \mathbf{y}(t) / \partial  \myvec{\bar{\psi}} = [[\partial \mathbf{y}(t) / \partial  \myvec{\bar{\psi}}_1]^\top,[\partial \mathbf{y}(t) / \partial\myvec{\bar{\psi}}_2]^\top \ldots [\partial \mathbf{y}(t) / \partial  \myvec{\bar{\psi}}_M]^\top]^\top$ with:
\begin{equation}\label{eq:phi-gradient-sim}
\footnotesize
\frac{\partial \mathbf{y}(t)}{\partial  \myvec{\bar{\psi}_m}} =
\begin{cases}
    \begin{array}{l}
        (\myvec{s}^{\top}(t)\mathbf{H}_{\rm 1}^{\ast}(t))  \\
        \otimes(\mathbf{H}_{\rm 2}(t) \prod_{m'=M}^{2} \myvec{\bar{\Psi}}_{m'} \myvec{\Xi}_{m'}\big)\myvec{D}, 
    \end{array} & \!\! \!\!m=1 \\[2ex]
    \begin{array}{l}
        \big(  \big(\prod_{m'=m}^{2} \myvec{\Xi}_{m'} \myvec{\bar{\Psi}}_{m'-1}  \big) \mathbf{H}_{\rm 1}^{\dagger}(t)\myvec{s}(t))^\top  \\
        \otimes(\mathbf{H}_{\rm 2}(t) \prod_{m'=M}^{m+1} \myvec{\bar{\Psi}}_{m'} \myvec{\Xi}_{m'} \big)\myvec{D}, 
    \end{array} & \!\! \!\!m=2,\dots,M
\end{cases}.
\end{equation}

\subsection{System Considerations During MINN Deployment}
Once our \gls{MINN} architecture is sufficiently trained, inference may take place on real data. This process can be also largely described by Algorithm~\ref{alg:SGD} with the omission of sampling $\myvec{o}(t)$ at line~$4$, since it is not available during inference, and, consecutively, the omission of line~$10$, where the backward pass is performed. In practice, it is reasonable to assume that training takes place at a single network node with sufficient computational and power capabilities, before sharing the obtained neural network weights to the corresponding physical devices once prior the commencement of the inference stage. On the other hand, when each \gls{DNN} module is physically collocated with its corresponding device, training entails the considerable overhead of exchanging gradient values using a dedicated control channel with sufficient capacity, which may result in costly power consumption by the \gls{TX}, \gls{RX}, and \gls{MS} devices.
Nevertheless, this methodology provides the benefit of enhanced privacy:
The observed input data by the \gls{TX} need not be transmitted to the \gls{RX}; the target information is solely required for the gradient computations given in~\eqref{eq:backprop_d}.

For the channel-aware transceiver cases, the implied channel estimation, at every time step, should be designed in a way that all physical entities receive the same estimates, which entails tailored orchestration and signaling. When reconfigurable \gls{MS}s are deployed, their control module should be large enough to perform effective feature extraction and phase configuration control. This dual need for data receiving capabilities and local computation could be addressed by the technology of \glspl{HRIS}~\cite{10352433}. In fact, by incorporating a power source to feed active components, sensing-capable \glspl{MS} are able to perform channel estimation locally. If endowed with DNN-optimized computation units, the channel-estimation-phase-configuration process may take place truly autonomously without any additional network overheads, other than the usual pilot exchange messages that need to take place regardless of the type of \gls{MS} device. The idea of self-configured \glspl{HRIS} to implement \gls{MS} controllers becomes increasingly attractive when the variation of channel-agnostic transceivers is adopted. Under this choice, the \gls{MS} is the only network node tasked with channel coding, which takes place over-the-air simultaneously to the inference process with important implications in the hardware requirements 
for the other devices. The numerical evaluation of the next section provides encouraging results that such a deployment scenario may be facilitated under certain network conditions.

On the opposite direction, a fixed-configuration \gls{MS} does not need to observe any \gls{CSI} during operation, which greatly simplifies the system requirements, as well as the overall computational budget.
In fact, \glspl{MS} with fixed phase configurations can be manufactured to be completely passive, essentially reducing a portion of the power consumption of the considered \gls{E2E} \gls{DNN}.
However, the channel coefficients still need to be available during training to compute the gradient updates, as described in the previous section, which again motivates the use of an \gls{HRIS} for efficient \gls{CSI} collection.

\begin{figure}
    \centering
    \includegraphics[width=\linewidth]{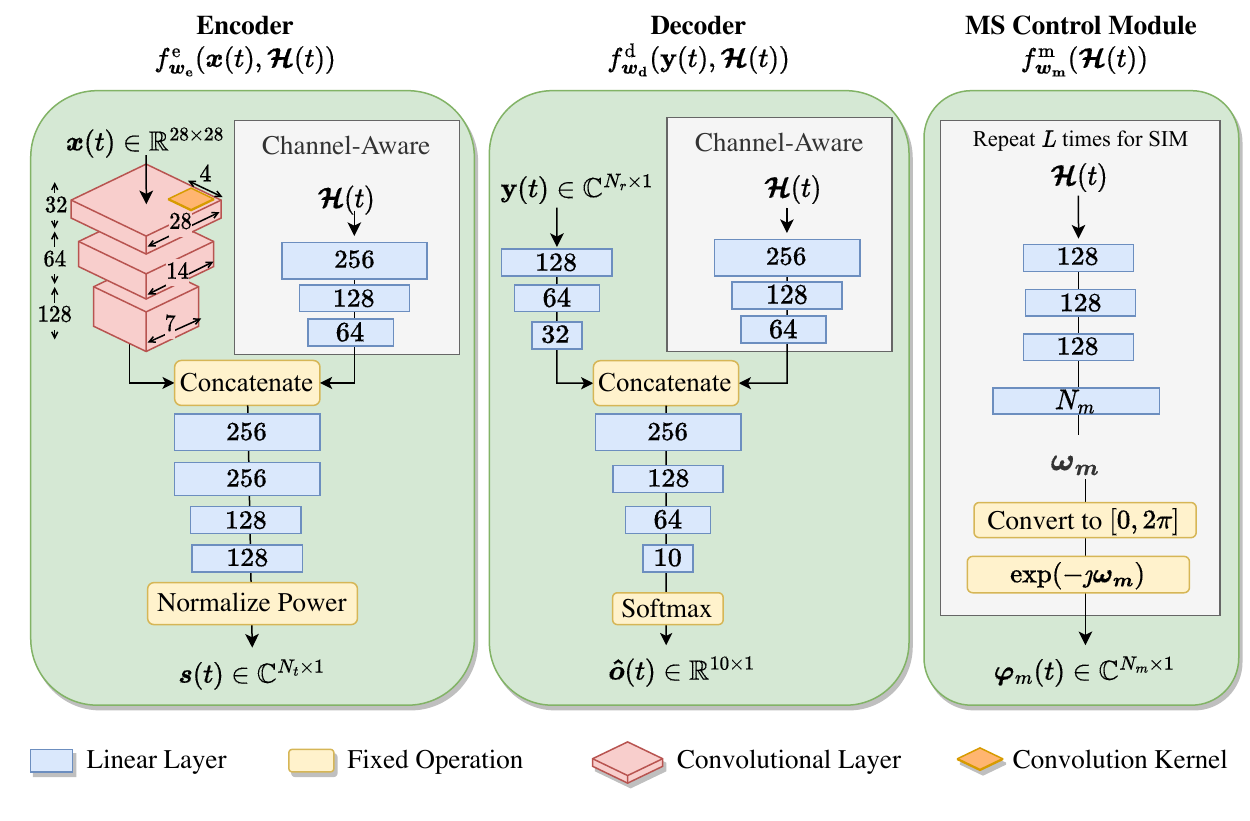}
    \vspace{-0.9cm}
    \caption{\gls{DNN} implementation of the three modules of the proposed \gls{MINN} architecture for the considered MNIST classification problem. The channel matrices of $\mathbf{\mathcal{H}}(t)$ are flattened to vectors and are concatenated. The channel-aware branches are ignored when channel-agnostic transceivers are used.}
    \label{fig:DNN-layers}
\end{figure}

\section{Numerical Results and Discussion}\label{sec:results}
\subsection{Simulation Setup}
To evaluate the proposed \gls{MINN} framework, we devise a problem of over-the-air MNIST classification~\cite{MNIST} under Ricean fading in the presence of an \gls{MS} in the wireless environment. We particularly consider an $N_t$-antenna low-cost \gls{IoT} \gls{TX} device  observing grayscale images of handwritten digits, and wishing to perform \gls{TOC} with an $32$-antenna \gls{RX}, which intends to obtain the numerical value of the digit of each image. A Cartesian coordinate system with the \gls{MS} at the origin has been simulated, where the \gls{TX} and \gls{RX} were placed at the points $(-2, 2, -0.5)$ and $(10,16,4)$, respectively. Narrowband transmissions at $28$~GHz with varying Ricean fading conditions were considered; specifically, we have set the Ricean factors to $13$, $7$, and $3$~dB for the \gls{TX}-\gls{MS}, \gls{MS}-\gls{RX}, and \gls{TX}-\gls{RX} channels, respectively. In fact, the details of the channel model follow the modeling of~\cite[Sec. 2]{ASH22}. It is noted that the \gls{TX}-\gls{RX} distance was approximately $19$~m, hence, the free-space attenuation of this direct link was at $41.5$~dB, while the total attenuation of the \gls{MS}-enabled \gls{E2E} link was approximately $67$~dB, indicating that, if the \gls{MS} was to be optimized to enhance the signal strength, only limited gains would be obtained due to the large pathloss difference. For the cases where the \gls{MS} was an \gls{RIS}, the surface was oriented parallel to the $xz$ plane to enable reflections, whereas, for the \gls{SIM} considerations, the cascaded surfaces had their broadside placed parallel to the $yz$ plane to enable diffraction from of signal arriving at their first layer, centered at the origin. Each of the next $M-1$ layers was placed in parallel at distances of $d_M = 5\lambda$ from its predecessor. The centers of the \gls{MS} elements had a distance of $\lambda/2$ with their neighbors, so that their surface area was $S_M=\lambda^2/4$. Finally, the transmission power was set to $P=30$~dBm, unless otherwise specified, with a noise variance of $\sigma^2=-90$~dBm, while the value $M=3$ for the \gls{SIM} layers was fixed.

\subsection{MINN Versions, Baselines, and Training Parameters}
The implementation details of the various simulated \gls{DNN} modules are included in Fig.~\ref{fig:DNN-layers}. Unless otherwise specified, ReLU activation functions were used, and layer-wise normalization was applied in all branches that received $\boldsymbol{\mathcal{H}}(t)$.
It is noted that, apart from the layers that are dependent on the system variables ($N_m$, $N_t$), the architectures of all \glspl{DNN} remained fixed for fairness in computational resources, meaning that, for larger parameter values, the networks could be less effective. The numbers of trainable parameters for all considered MINN versions are included in Table~\ref{tab:DNN-param-sizes}, illustrating the computational requirement increase in channel-aware settings and in case of larger \glspl{MS}.

The simulated MINN versions were compared against two baselines. The first one was a baseline of the same encoder and decoder modules, but in the absence of any \gls{MS}; this baseline enables the evaluation of the specific benefits brought by the \gls{MS}. In fact, this is accommodated by our system model, by setting $\myvec{\Omega}(t) = \boldsymbol{0}_{N_m\times N_m}$ in \eqref{eq:received-signal-RIS_1}.
The second baseline implements the ``transmit-then-infer'' paradigm as discussed in Section~\ref{sec:EI-theory}, referred to as ``RX-DNN,'' where the RX reconstructs the data and employs a \gls{DNN} classifier trained independently of the channel optimization phase.
It considers a conventional communication system, which is designed to encode, modulate, and apply TX/RX beamforming to the image data, upon optimizing the channel's data rate.
The details of the system are given in the Appendix.

The MNIST data set contains $7\times 10^4$ images, $10^4$ of which were used for testing. Additionally, $5\times10^3$ channel realizations were used for training, while $10^3$ more were used for testing. At each forward pass, during either training or inference, the input image was paired with a random channel realization. Each training instance took place over $200$ epochs, using the Adam optimizer~\cite{Adam} with learning rate of $\eta=10^{-4}$ and a weight decay of $10^{-4}$. It has been observed that this effective expansion of the data set to $3\times10^8$ image-channel pairs, as well as the stochastic nature of channel realizations, increase the variance of the achieved \gls{DNN} performance across multiple restarts of the training process. For all experiments, each method was trained $10$ times with different initialization seeds, and the top, median, first (Q1), and third (Q3) percentiles of the achieved test set accuracy scores are reported.


\begin{table}[t]
    \centering
    \caption{Number of DNN Parameters for Different MINNs and $N_t=4$.}
    \renewcommand{\arraystretch}{1.2}
    \setlength{\tabcolsep}{2.5pt} 
    \begin{tabular}{|p{2.1cm}@{}|c|c|c|}
        \hline
        \textbf{Method (MS} & \textbf{Controllable MS} & \textbf{Fixed MS \&} & \textbf{Fixed MS \&} \\
        \textbf{elements)}& \textbf{\& Channel-Aware} & \textbf{Channel-Aware} & \textbf{Channel-Agnostic} \\
        \hline
        SIM ($3\!\times\!8\!\times\!8$)   & $2.3\!\times\!10^6$ & $2.3\!\times\!10^6$ & $2.5\!\times\!10^5$ \\
        SIM ($3\!\times\!12\!\times\!12$) & $4.6\!\times\!10^6$ & $4.6\!\times\!10^6$ & $2.5\!\times\!10^5$ \\
        RIS ($16\!\times\!16$)            & $1.0\!\times\!10^7$ & $7.7\!\times\!10^6$ & $2.5\!\times\!10^5$ \\
        RIS ($25\!\times\!25$)           & $2.5\!\times\!10^7$ & $1.8\!\times\!10^7$ & $2.5\!\times\!10^5$ \\
        \hline
        No Metasurface & \multicolumn{2}{c|}{$9.9\!\times\!10^5$} & $2.5\!\times\!10^5$ \\
        \hline
        RX-DNN ($25\!\times\!25$) & \multicolumn{3}{c|}{$1.3\!\times\!10^6$} \\
        \hline
    \end{tabular}
    \label{tab:DNN-param-sizes}
\end{table}

\subsection{Reconfigurable versus Fixed Metasurfaces}

\begin{figure}[t]
    \centering
    \includegraphics[scale=0.45]{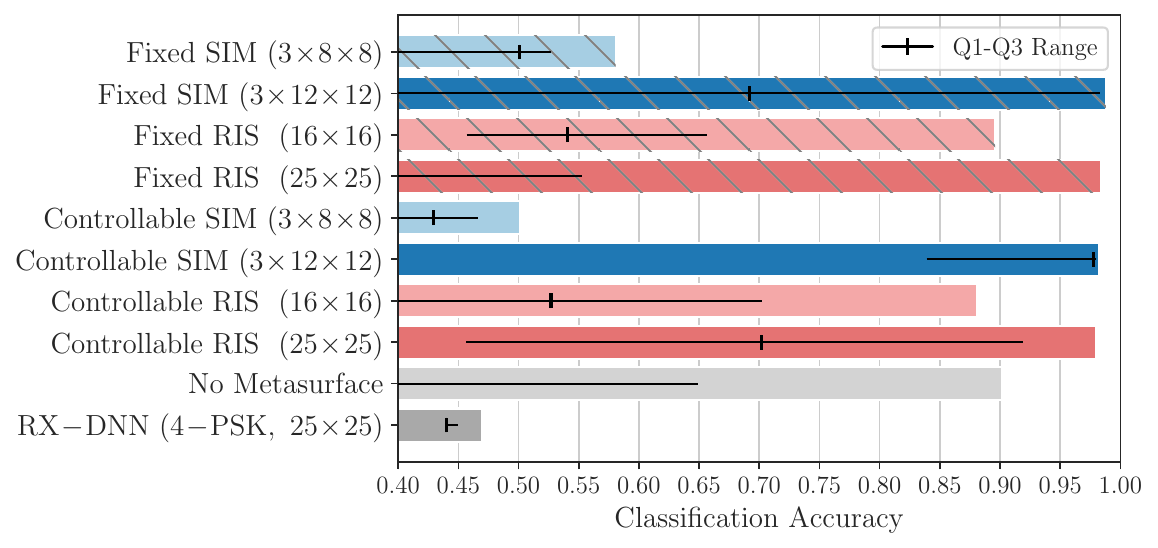}
    \caption{Comparison of achieved accuracy with different \gls{MINN} variations and the two adopted baselines, considering $N_t=4$ and channel-aware transceivers. Bars indicate the highest performance with each method across multiple training restarts, while the black horizontal lines represent the Q1-Q3 range, with each black vertical line representing the median performance.}
    \label{fig:reconfigurable-vs-static-Nt2}
\end{figure}

\begin{figure}[t]
    \centering
    \includegraphics[width=\linewidth]{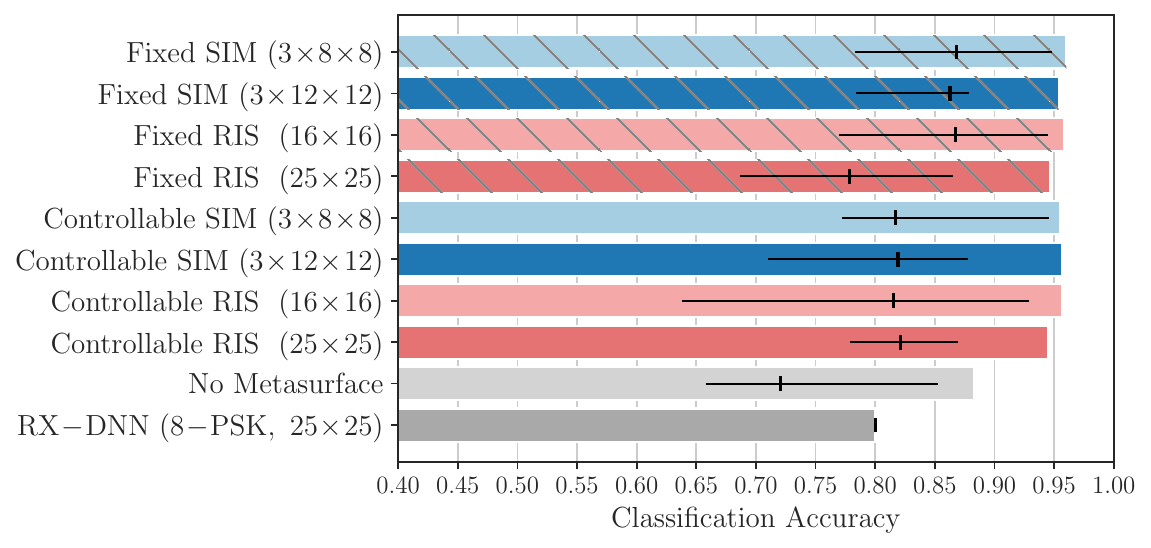}
    \caption{Similar to Fig.~\ref{fig:reconfigurable-vs-static-Nt2} but for $N_t=12$.}
    \label{fig:reconfigurable-vs-static-Nt12}
\end{figure}
We commence by investigating the benefits of \glspl{MINN} incorporating reconfigurable \glspl{MS} (that entail an extra \gls{DNN} module at the \gls{MS} controller) in comparison with their trainable fixed-MS-response counterparts (labeled as ``Fixed'' in the sequel), considering channel-aware transceivers. The test set classification accuracy considering two different \gls{RIS} and \gls{SIM} sizes and a TX with $N_t=4$ is illustrated in Fig.~\ref{fig:reconfigurable-vs-static-Nt2}. As shown, the \gls{MINN} versions with larger \gls{RIS}/\gls{SIM} sizes achieve near optimal performance, however, smaller sizes, especially for the cases of \glspl{SIM}, are less effective. It is also depicted that, in the absence of an \gls{MS}, substantial accuracy is reached.
However, the classification task is not solved successfully, since the classification accuracy on MNIST data of even simple \gls{DNN} classifiers reach scores higher than $0.98$ (see also the Appendix).
It is also evident that the baseline communication system performs poorly due to difficulties in correctly detecting the transmitted symbols.
Few antenna elements and \gls{LoS}-dominant channels restrict the number of available data streams, $d$, in spatial multiplexing, resulting to denser constellations.

Figure~\ref{fig:reconfigurable-vs-static-Nt12}, where the number of TX antennas is increased to $N_t=12$, demonstrates that all considered \gls{MINN} variations are successful in their \gls{TOC} task, in contrast to the two baselines. Notice that the variability across trials, indicated by the Q1-Q3 ranges, is much reduced compared to the $N_t=4$ case in Fig.~\ref{fig:reconfigurable-vs-static-Nt2}, indicating that, as expected, \gls{MINN} approaches become more robust as the communication resources increase. In addition, by inspecting  Figs.~\ref{fig:reconfigurable-vs-static-Nt2} and~\ref{fig:reconfigurable-vs-static-Nt12}, it can be seen that  there is no observed difference in the performance between the controllable and fixed versions of all \gls{MINN} versions. It can, thus, be concluded that fixed-MS-response approaches provide {\em more effective E2E DNN architectures} in static wireless systems, in view of the reduced computational, hardware, and system requirements, as discussed in Section~\ref{sec:training-deployment}.

\subsection{Comparison with Channel-Agnostic Schemes}
\begin{figure}[t]
    \centering
    \includegraphics[width=\linewidth]{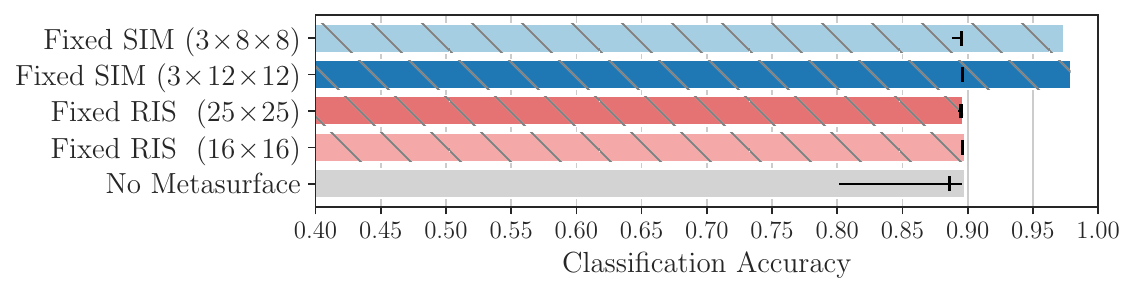}
    \caption{Achieved accuracy with different \gls{MINN} variations and the two adopted baselines, considering $N_t=6$ and channel-agnostic transceivers.}
    \label{fig:results-no-CSI}
\end{figure}
\gls{MINN} variations with fixed-MS-responses, channel-agnostic transceivers, and $N_t=6$ are compared in Fig.~\ref{fig:results-no-CSI}. 
It can be observed that both \gls{SIM}-based architectures achieve {\em near optimal performance without the need of \gls{CSI} acquisition} during network deployment, which has important benefits in the simplification of the overall \gls{TOC} system design. Interestingly, when \glspl{RIS} are deployed, the performance gets degraded to equal levels as the no-\gls{MS} baseline. It is noted that the absence of \gls{CSI} leads to smaller E2E \gls{DNN} sizes, due to the exclusion of the channel coding branches whose parameters grow linearly with the number of channel coefficients.
This reduction in \gls{DNN} size correlates with the reduced variability of the methods' performances across re-runs, which hints that much more extended training periods would be required for effective training the channel-aware architectures of the previous section.



\subsection{The Role of the Transmission Power during Training}

\begin{figure}[t]
    \centering
    \includegraphics[width=0.9\linewidth]{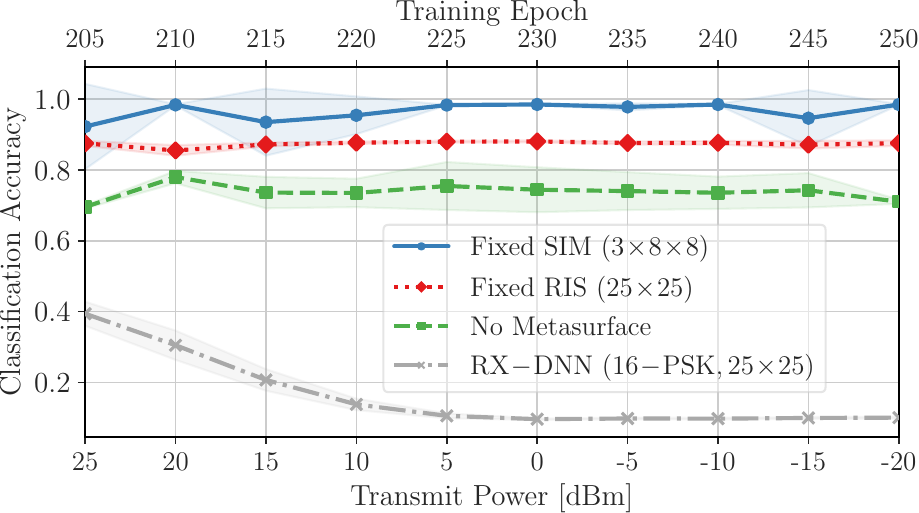}
    \caption{Achieved accuracy with fixed-configuration \glspl{MINN} and the two adopted baselines, considering channel-agnostic transceivers, for progressive TX power decrement over training.}    
    \label{fig:results-varying-P}
\end{figure}
Trained \glspl{DNN} are expected to be robust to increased noise levels in the inference data sets without or with minimal extra training. To investigate this hypothesis, we have implemented the following procedure for two \gls{MINN} versions with fixed-MS-responses, considering channel-agnostic transceivers. Once the first $200$ training epochs elapsed, we continued the training process by decreasing the transmission power by $5$~dBm every five epochs, starting from $P=30$~dBm until we reach $-20$~dBm. In this way, the first epochs were used as pre-training, while the latter ones fine-tuned the model to the changing conditions. The motivation for this approach is that \gls{DNN}s typically perform most of their learning during the initial iterations, while few final iterations only refine the learning process by imposing small changes to the learned weights. Recall that, for the classification performance results in Figs.~\ref{fig:reconfigurable-vs-static-Nt2}--\ref{fig:results-no-CSI}, $P=30$~dBm was used throughout the entire training process to provide high \gls{SNR} conditions.

As illustrated in Fig.~\ref{fig:results-varying-P}, all considered \gls{MINN} approaches, as well as the baseline scheme without an \gls{MS}, exhibit stable performance despite the power drop. It can be seen that the fixed-\gls{SIM}-response \gls{MINN} approach provides the best classification accuracy, indicating that once trained, it can successfully solve the inference task {\em even at $\mathit{50}$~dB of lower \gls{SNR} levels}, which offers substantial energy benefits in the deployment of such \gls{TOC} approaches. Finally, as expected, the performance of the traditional communication system is heavily impacted with the decrease of the transmission power.

\section{Conclusions}\label{sec:conclusion}
In this paper, we proposed the framework of \glspl{MINN} that enables \gls{EI} by treating the \gls{MS}-programmable wireless channel as a hidden \gls{OAC} artificial neural network layer(s), in sharp contrast to previous literature that considers the signal propagation environment as a source of noise. Architectural \gls{MINN} variations have been presented that consider \glspl{RIS} and \gls{SIM}, either controllable via dedicated \gls{DNN} modules or with trainable fixed response configurations, while both channel-aware and channel-agnostic transceivers have been considered.
A variation of the backpropagation algorithm for fading channels has been developed, and system considerations that concern data collection, channel acquisition, and hardware requirements have been discussed.
Our performance evaluations highlighted that \glspl{MINN} are more efficient classification schemes compared to \gls{MS}-free or traditional ``transmit-then-compute'' systems, with the same link budget measured in terms of TX power, number of antennas and \gls{MS} elements.
The key insights of our numerical analysis showcased that, for the considered system parameters, fixed \gls{MS} responses are more effective \gls{E2E} \gls{DNN} architectures in view of the reduced computations, hardware, and system requirements, while \gls{CSI} knowledge is not a hard requirement for the transceiver, since channel coding may take place over-the-air via \gls{SIM}. Finally, it was demonstrated that, once pre-trained under high \gls{SNR}, fixed-MS-response \gls{MINN} architectures can be sufficiently robust to a wide range of \gls{SNR} conditions, offering a strong comparative advantage for practical \gls{EI} deployment. 

\appendices

\begin{figure}
    \centering
    \includegraphics[width=\linewidth]{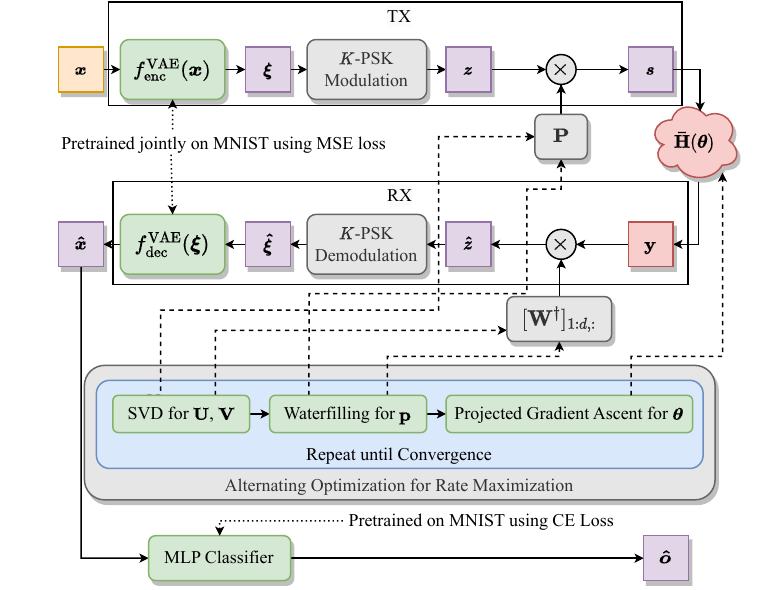}
    \caption{Block diagram of a conventional MIMO communication system for MNIST classification with independent source and channel coding.}
    \label{fig:baseline-architecture}
\end{figure}

\section*{Appendix \\Baseline Communication System for MNIST} \label{sec:appendix}

In this section, we design a MIMO communication system for MNIST classification with independent source and channel coding. For ease of notation, we drop index $t$, noting that all operations take place at every channel frame.
First, a \gls{VAE}~\cite{VAE} is used as a data-driven lossy compression scheme, whose operation is defined as $\boldsymbol{\hat{x}}\triangleq f_{\rm dec}^{\rm VAE}(f_{\rm enc}^{\rm VAE}(\boldsymbol{x}))$, where the encoder (that is placed at the \gls{TX}) generates a latent space representation $\boldsymbol{\xi} \triangleq f_{\rm enc}^{\rm VAE}(\boldsymbol{x})\in\mathbb{R}^{2\times1}$, while the decoder (at the \gls{RX} side) learns to reconstruct the original input $\boldsymbol{x}$ as $\boldsymbol{\hat{x}}=f_{\rm dec}^{\rm VAE}(\boldsymbol{\xi})$.
The \gls{DNN} architecture for \gls{VAE} is kept as in~\cite{VAE}, achieving $O(10^{-4})$ \gls{MSE} reconstruction error once pretrained on MNIST data. The two elements of $\boldsymbol{\xi}$, being, for example, two $32$-bit floating point numbers, are then modulated via $K$-ary \gls{PSK}, where $K$ is determined empirically to obtain the number of transmitted symbols $d=\lceil 64/ {\rm log}_2 K \rceil \leq \min\{N_t,N_r\}$, which are included in the symbol $\boldsymbol{z}  \in \mathbb{C}^{d \times 1}$, as shown in Fig.~\eqref{fig:baseline-architecture}.

By denoting the \gls{RIS}-aided \gls{E2E} channel as $\mathbf{\bar{H}}(\boldsymbol{\theta}) \triangleq \mathbf{H}_{\rm D}+\mathbf{H}_{\rm 2} {\rm diag}(\exp(-\jmath \myvec{\theta})) \mathbf{H}_{\rm 1}^{\dagger}$ and its \gls{SVD} as $\mathbf{\bar{H}}(\boldsymbol{\theta}) = \mathbf{U}\mathbf{\Sigma}\mathbf{V}^\dagger$, $\mathbf{P} \triangleq {\rm diag}(\sqrt{\myvec{p}}) [\mathbf{V}]_{:,1:d}\in \mathbb{C}^{N_t \times d}$ can be used as the precoding matrix with $\myvec{p}$ including the per-symbol power allocation, i.e., the transmitted signal is $\myvec{s}  = \mathbf{P} \boldsymbol{z} $.
At the \gls{RX}, weighted minimum mean squared error combining is adopted, yielding the estimation $\boldsymbol{\hat{z}}  = [\mathbf{W}^\dagger]_{1:d,:} \mathbf{y}$ with $\mathbf{W} \triangleq \left( \bar{\mathbf{H}}(\boldsymbol{\theta}) \mathbf{P} \mathbf{P}^\dagger \bar{\mathbf{H}}^\dagger(\boldsymbol{\theta}) + \sigma^2 \mathbf{I} \right)^{-1} \bar{\mathbf{H}}(\boldsymbol{\theta}) \mathbf{P}$.
Notations $[\mathbf{V}]_{:,1:d}$ and $[\mathbf{W}^\dagger]_{1:d,:}$ represent respectively the first $d$ columns of $\mathbf{V}$ and the first $d$ rows of $\mathbf{W}^\dagger$.
Then, the rate maximizing power allocation $\boldsymbol{p}$ and the RIS phase configuration $\boldsymbol{\theta}$ can be obtained from the solution of the optimization problem:
\begin{align*}
    \mathcal{OP}_1:\,\, &\underset{\boldsymbol{p}, \boldsymbol{\theta}}{\max}\sum_{i=1}^{\min\{N_t, N_r\}} \log_2\left(1+\frac{[\boldsymbol{p}]_i \tilde{\sigma}_i(\boldsymbol{\theta}) }{\sigma^2} \right) \\
    &\ \mathrm{s.t.}\,\, \| \boldsymbol{p}\|=P,\, {\theta}_n \in [0, 2\pi) \,\, \forall n=1,2,\dots,N_m,
\end{align*}
where $\tilde{\sigma}_i(\boldsymbol{\theta})$ is the $n$-th singular value of $\mathbf{\bar{H}}(\boldsymbol{\theta})$. The latter approach can be followed in an alternating optimization manner to design the TX precoder, RX combiner, and RIS phase configuration maximizing the achievable spectral efficiency. In particular, we iterate until convergence between: Step~1) for a given $\boldsymbol{\theta}$, obtain $\boldsymbol{p}$ from the waterfilling solution of $\mathcal{OP}_1$ as well as $\mathbf{P}$ and $\mathbf{W}$ using also $\mathbf{\bar{H}}(\boldsymbol{\theta})$'s SVD; and Step~2) for previous step's $\boldsymbol{p}$ as well as $\mathbf{V}$ and $\mathbf{U}$ from the previous iteration, perform projected gradient ascent using automatic differentiation to find $\boldsymbol{\theta}$ solving $\mathcal{OP}_1$.

Once $\boldsymbol{\hat{z}} $ is retrieved, it is demodulated to obtain $\boldsymbol{\hat{\xi}}$, which is then fed into $f_{\rm dec}^{\rm VAE}(\boldsymbol{\hat{\xi}})$ to obtain the reconstructed image $\boldsymbol{\hat{x}}$. A \gls{MLP} classifier, complementing our baseline communication system summarized in Fig.~\eqref{fig:baseline-architecture}, has been pretrained on original MNIST images (i.e., instances of $\boldsymbol{x}$) with over $0.98$ classification accuracy. In our experimentation, we have evaluated \gls{MLP} on unseen instances of $\boldsymbol{\hat{x}}$ to assess this baseline's classification accuracy.


\FloatBarrier
\bibliographystyle{IEEEtran}
\vspace{-0.1cm}
\bibliography{references}

\end{document}